\PassOptionsToPackage{table}{xcolor}
\documentclass{article}

 \usepackage[preprint]{neurips_2026}

\usepackage[utf8]{inputenc} 
\usepackage[T1]{fontenc}    
\usepackage{hyperref}       
\usepackage{url}            
\usepackage{booktabs}       
\usepackage{amsfonts}       
\usepackage{nicefrac}       
\usepackage{microtype}      
\usepackage{xcolor}         
\usepackage{booktabs}
\usepackage{multirow}
\usepackage{tabularx}
\usepackage{array}
\usepackage{makecell}
\renewcommand{\arraystretch}{1.12}
\usepackage{booktabs}
\usepackage{xcolor}
\usepackage{longtable}
\usepackage{array}
\usepackage{multirow}
\usepackage[table]{xcolor}
\usepackage{graphicx}
\usepackage{float}
\usepackage{amsmath}
\usepackage{subcaption}
\usepackage{adjustbox}
\usepackage{array}
\newcolumntype{L}[1]{>{\raggedright\arraybackslash}m{#1}}
\newcolumntype{C}[1]{>{\centering\arraybackslash}m{#1}}

\newcommand{\high}{\textcolor{green!50!black}{High}}
\newcommand{\medium}{\textcolor{orange!85!black}{Medium}}
\newcommand{\low}{\textcolor{red!70!black}{Low}}

\definecolor{groupgray}{gray}{0.94}

\title{Unpacking the Evaluators: How Configuration Shapes the Evaluation of Alignment in Explainable AI}

\author{%
  Gizem Karagoz \\
  Eindhoven University of Technology \\
  \texttt{g.karagoz@tue.nl}
  \And
  Tanir Ozcelebi \\
  Eindhoven University of Technology \\
  \texttt{t.ozcelebi@tue.nl}
  \And
  Nirvana Meratnia \\
  Eindhoven University of Technology \\
  \texttt{n.meratnia@tue.nl}
}

\begin{document}

\maketitle

\begin{abstract}

Explainable AI (XAI) evaluation fundamentally requires human-centered evaluation, assessing whether explanations align with human reasoning and expert knowledge. However, such evaluations often require expert-driven manual validation and are difficult to scale, leading to the widespread use of alignment-based metrics, where attribution maps are compared against predefined ground-truth (GT) representations. Despite their popularity, alignment evaluations are implicitly bias towards multiple design choices made during the evaluation process, for example GT representations, metric families, metric formulations (soft vs.\ hard), and thresholding strategies. In this paper, we formalize alignment evaluation as a structured evaluation configuration space through which the influence of these design choices can be systematically analyzed. We further introduce a structured alignment evaluation framework that models interactions between XAI methods, GT representations, metric families, metric formulations, and thresholding strategies across 2D and 3D synthetic and real-world datasets. In addition, we propose a novel taxonomy of XAI output behavior based on local entropy and boundary mass ratio, enabling structured characterization of attribution patterns for alignment evaluation. 
Our experiments reveal systematic flows (triples of XAI-GT-metric configurations), showing that alignment outcomes provide distinct behavior patterns under certain configurations, such as soft/hard metric implementations.  We further observe varying levels of threshold sensitivity across metric families and attribution structures. These findings demonstrate that alignment evaluation cannot be treated as a fixed or invariant measure of explanation quality, and provide practical guidance for designing more reliable and behavior-aware XAI evaluation protocols.

\end{abstract}

\section{Introduction}

Explainable AI (XAI) aims to make machine learning models more transparent by providing explanations for their predictions \citep{miller2019explanation}. In high-stake domains such as healthcare, these explanations must ultimately be validated from a human-centered perspective, assessing whether they align with expert reasoning and support reliable decision-making~\citep{tonekaboni2019clinicians, amann2020explainability, ghassemi2021false}. However, human-centered evaluation requires expert-driven studies and repeated manual validation, which makes the process time-consuming and difficult to scale across models, datasets, and explanation methods~\citep{rong2023towards, nauta2023anecdotal}.

To address this challenge, alignment evaluation has emerged as a practical proxy \citep{doshi2017towards}. These approaches compare explanation maps with predefined ground-truth (GT) representations, such as segmentation masks, bounding boxes, or key points, to quantify how well explanations correspond to expected evidence~\citep{selvaraju2017grad, petsiuk2018rise, deyoung2020eraser}. Such approaches are therefore widely adopted in computer vision and medical imaging settings where expert annotations are available.

Despite their widespread adoption, alignment-based evaluations rely on multiple design choices. One important factor is the representation of the ground truth itself. Existing studies commonly evaluate explanations using segmentation masks, bounding boxes, or point annotations~\citep{selvaraju2017grad, petsiuk2018rise, deyoung2020eraser}, although these representations capture different notions of evidence, such as region coverage, structural boundaries, or sparse localization indicators~\citep{cheng2021boundary}. 

Evaluation outcomes also depend on the choice of alignment metric. Overlap-based metrics emphasize spatial agreement~\citep{rezatofighi2019generalized}, point-based metrics focus on peak localization~\citep{zhang2018top}, mass-based metrics quantify attribution concentration~\citep{wang2020score}, while distance-based metrics capture structural discrepancy~\citep{huttenlocher2002comparing}. Prior studies have shown that different metrics can produce inconsistent rankings and conflicting conclusions regarding explanation quality~\citep{hooker2019benchmark, klein2024navigating}.

In addition, alignment metrics can be computed either on continuous magnitude-aware attribution maps (soft evaluation) or on thresholded discrete binary supports (hard evaluation). As highlighted by \citep{chase2024attribution}, discrete evaluation protocols typically rely on top-\(k\) thresholding, which introduces threshold selection bias and extreme sensitivity to attribution sparsity. 

XAI methods are classified under diferent taxonomies (i,e, post-hoc/ante-hoc, local/global, attribution-based, gradient-based)~\citep{arrieta2020explainable, deng2026attribution, speith2022review} but not considering their spatial output behaviors. However, attribution methods exhibit diverse spatial structures, from localized and compact to diffuse or scattered patterns. These variations significantly skew evaluation outcomes; for instance, methods acting as edge detectors can achieve high alignment scores despite being insensitive to model parameters \citep{adebayo2018sanity}.

In this paper, we study alignment evaluation itself as an object of analysis, covering all the above-mentioned design choices. We conduct a systematic analysis across 2D and 3D synthetic and real-world datasets, considering multiple GT representations, diverse metric families, and both soft and hard evaluation settings. To capture interactions between these components, we model alignment as flows (triples across XAI methods, GT representations, and metric families as the configuration). In addition, we characterize explanation behavior using local entropy and boundary mass ratio, enabling to identify the output behavior of attribution maps for defining a proper alignment configuration for evaluation. For further analysis, threshold sensitivity is analyzed to assess the stability of evaluation results.


\textbf{Our contributions are as follows:}
\begin{itemize}
    \item We formalize alignment evaluation as a configuration space, through which the impact of XAI methods, ground-truth (GT) representations, metric families, metric formulations, and thresholding on evaluation outcome can be systematically analyzed.

    \item We introduce a structured alignment evaluation framework that models interactions between XAI methods, GT representations, and metric families under both soft and hard formulations, enabling principled analysis of evaluation compatibility across configurations.

    \item To the best of our knowledge, we propose the first taxonomy of XAI output behavior based on local entropy and boundary mass ratio, enabling structured characterization of explanation maps for alignment evaluation and broader XAI analysis.

    \item Through extensive experiments across synthetic and real-world datasets, we identify consistent relationships between XAI methods, GT representations, metric families, and thresholding behavior, which provides practical guidance for selecting and interpreting alignment evaluation configurations in future studies.
\end{itemize}

\section{Related Works}

\paragraph{Metric selection.}
Alignment-based XAI evaluation commonly relies on localization metrics such as Intersection over Union (IoU), Pointing Game (PG), and energy-based variants to compare attribution maps with GT as reference regions~\citep{le2023benchmarking, selvaraju2017grad, petsiuk2018rise, zhang2018top}. However, these metrics capture different notions of agreement. For example IoU emphasizes spatial overlap, PG focuses on peak localization, and energy-based metrics account for attribution mass. Prior work has shown that XAI evaluation can yield inconsistent outcomes in terms of accuracy, depending on the chosen metric and evaluation protocol~\citep{hooker2019benchmark}. Additionally, recent benchmarking work has emphasized that design choices and metric selection can lead to conflicting rankings of XAI methods~\citep{klein2024navigating}. 

\paragraph{Ground-truth representation.}
Alignment evaluation also depends on how the ground truth is represented. Mullen et al. \citep{mullen2019comparing} analyze the tradeoffs in cost and accuracy among three spatial ground truth formats: polygons, bounding boxes, and centroids. The study finds that while polygons are the most accurate, bounding boxes and centroids are significantly more efficient to produce and often provide similar performance in machine learning detection tasks. In another study\citep{agnew2023quantifying}, the selection of ground truth type between polygon(i.e. region) and bounding box has changed the segmentation outcomes. Cheng et al.~\citep{cheng2021boundary} demonstrate that standard Mask IoU is biased toward larger objects and often fails to capture poor boundary alignment, as interior pixels dominate the score.

\paragraph{XAI Alignment Evaluation.}
 Grad-CAM~\citep{selvaraju2017grad} evaluates localization performance using bounding-box annotations and weak localization criteria, while RISE~\citep{petsiuk2018rise} and Score-CAM~\citep{wang2020score} employ Pointing Game, deletion/insertion, and energy-based metrics over saliency maps and region-based GT. CLEVR-XAI~\citep{arras2022clevr} further introduces pixel-level evaluation protocols based on relevance mass and relevance rank metrics computed directly on continuous attribution maps. More recent benchmarking frameworks, such as QUANTUS~\citep{hedstrom2023quantus} and the large-scale analysis of Klein et al.~\citep{klein2024navigating}, provide broad collections of XAI evaluation metrics and benchmark protocols across methods and datasets.

Prior work has highlighted important challenges in alignment-based XAI evaluation, including metric inconsistency problems, variability across GT representations and its effect on the final outcome, and sensitivity to thresholding choices. Existing studies typically rely on fixed evaluation settings, while GT representation, metric family, evaluation formulation, and threshold sensitivity are either treated independently or remain implicitly defined or unexplored. In this work, we formalize alignment evaluation as a structured evaluation space that enables systematic analysis of how these design choices shape evaluation conclusions.

\section{Alignment Evaluation Framework}
\label{sec:methodology}
\begin{figure}[h!]
    \centering
    \includegraphics[width=\textwidth]{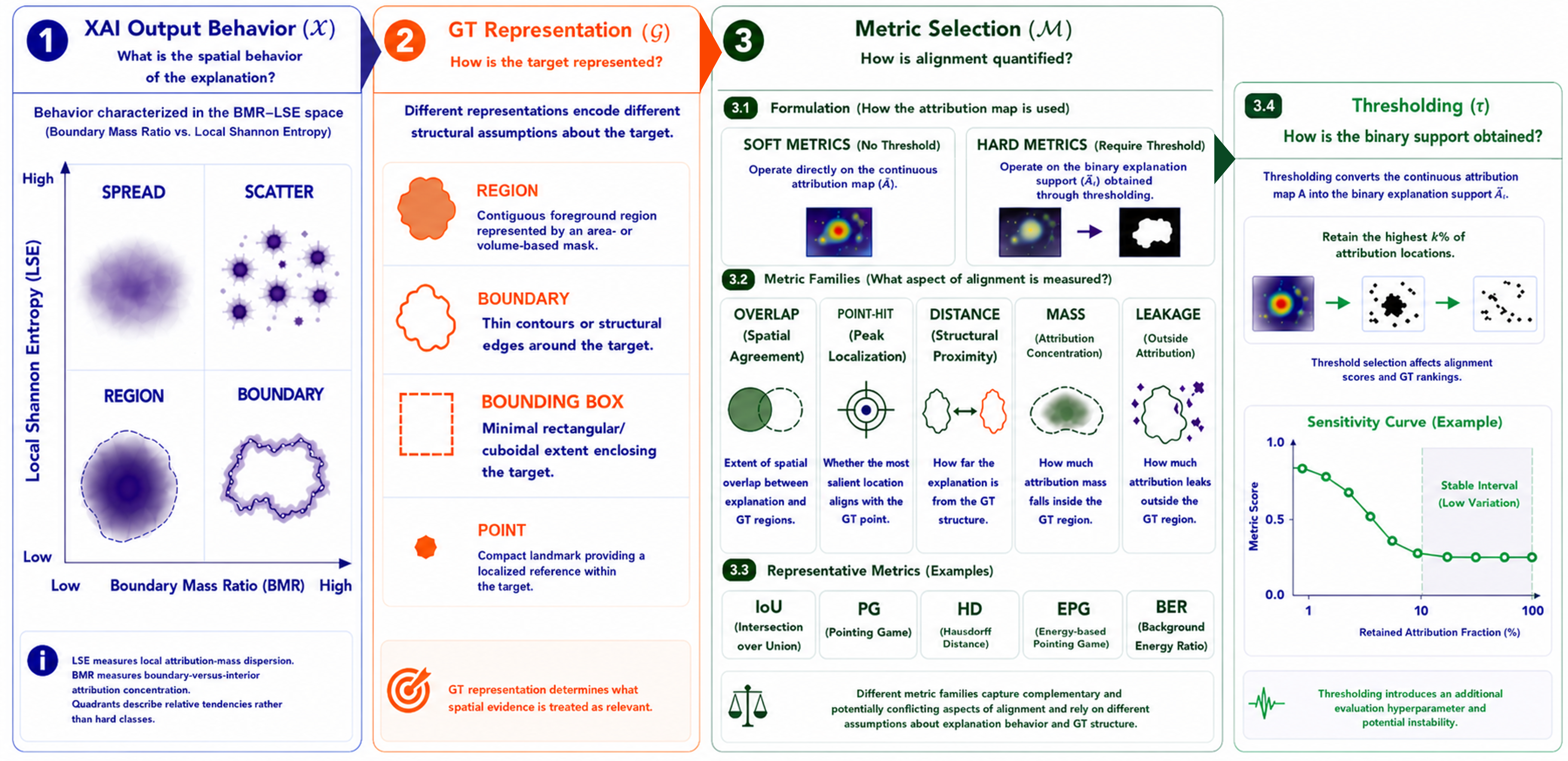}
    \caption{Alignment outcomes are determined by three interacting components: (1) XAI output behavior characterized in the LE–BMR space, (2) ground-truth (GT) representation, (3) metric selection, including metric families and soft vs. hard formulations, and (3.4) thresholding for hard metrics. The figure illustrates how different configurations of these components lead to variations in alignment scores and rankings.}
    \label{fig:methodology}
\end{figure}

Figure~\ref{fig:methodology} illustrates our alignment evaluation framework, driven by four interacting components: XAI output behavior, GT representation, metric selection, thresholding of attribution maps for hard metric implementations, and the  interactions between these components.

\subsection{Evaluation Configuration Space}

We formulate XAI alignment evaluation as a configuration-dependent process rather than as a fixed comparison between an explanation and a ground truth only as it is commonly done in the existing literature. Given an XAI method $e$, a ground-truth representation $g$, an alignment metric $m$, evaluation formulation(soft/hard) $v$, and threshold parameter $\tau$ (only for hard metrics), we define an evaluation configuration as
\[
\mathcal{C} = (e, g, m, v, \tau),
\]

Given an input sample $x$ and predictive model $f$, the resulting alignment score under configuration $\mathcal{C}$ is defined as
\[
S_{\mathcal{C}} =
m\big(A_e(x,f), G_g(x)\big),
\qquad
S_{\mathcal{C},\tau} =
m\big(\tilde{A}_{e,\tau}(x,f), G_g(x)\big).
\]
where $A_e(x,f)$ denotes the attribution map produced by explanation method $e$, $\tilde{A}_{e,\tau}(x,f)$ denotes its thresholded support under threshold $\tau$, and $G_g(x)$ denotes the GT representation used for evaluation.

This formulation defines alignment evaluation as a structured measurement configuration rather than a fixed evaluation protocol.

\subsection{Characterization of XAI Output Behavior}

Attribution maps produced by different XAI methods exhibit diverse spatial structures and concentration patterns, which can influence alignment evaluation outcomes under different GT representations and metric families. To systematically characterize these attribution behaviors, we used two complementary measures: boundary mass ratio (BMR) and local entropy (LE).

\paragraph{Boundary Mass Ratio (BMR).}
measures proportion of attribution mass concentrated along the boundary of the explanation support relative to its total attribution mass. Let $A$ denote an attribution map and $S(A)$ its retained attribution
support. The inner morphological boundary of this support is denoted by
$\partial S(A)$. BMR measures the proportion of retained attribution
mass located on this boundary:
\[
\operatorname{BMR}(A)=
\frac{\sum_{i\in\partial S(A)}|A_i|}
     {\sum_{i\in S(A)}|A_i|}.
\]
For non-zero retained attribution mass, $\operatorname{BMR}(A)\in[0,1]$.
Higher values indicate stronger attribution-mass concentration on the
support boundary, whereas lower values indicate greater concentration
in its interior. BMR is GT-independent because both $S(A)$ and
$\partial S(A)$ are derived exclusively from the attribution map.

\paragraph{Local Entropy (LE).}
captures spatial coherence of attribution values. We compute patch-wise Shannon entropy over the attribution map and average across patches \citep{wu2013local}. The retained attribution map is partitioned into $K$ non-overlapping spatial patches $\Omega_k$. For the $n_k>1$ active locations in patch $\Omega_k$, attribution magnitudes are normalized as $\pi_{k,i}=|A_i|/\sum_{j\in\Omega_k\cap S(A)}|A_j|$. The normalized patch entropy and LE are
\[
H(\Omega_k)=
-\frac{\sum_i\pi_{k,i}\log\pi_{k,i}}{\log n_k},
\qquad
\operatorname{LE}(A)=\frac{1}{K}\sum_{k=1}^{K}H(\Omega_k).
\]
Empty patches and patches with one active location are assigned
$H(\Omega_k)=0$ and remain in the final mean. Thus, $\operatorname{LE}(A)\in[0,1]$. Lower entropy indicates structured and coherent patterns, while higher entropy indicates scattered or noise-like distributions. 


\paragraph{Behavior Types.}
Combining BMR and local entropy (LE), we identify four  behavior types (Fig.~\ref{fig:methodology}). Region patterns correspond to compact and coherent attributions, boundary patterns emphasize edge-focused responses, spread patterns reflect diffuse attribution over broader areas, and scatter patterns exhibit fragmented and noisy distributions.

We apply this characterization consistently to both 2D and 3D attribution maps. LE and BMR together provide complementary measurements of attribution coherence and structural concentration, enabling the proper analysis of alignment evaluation.

\subsection{Ground Truth (GT) Representation}

Evaluation of explanation methods requires a GT reference against which attribution maps are compared. However, GT is not uniquely defined: the same underlying target can be represented in multiple ways, such as region masks, boundaries, bounding boxes, or point annotations~\citep{cheng2021boundary}. 

\begin{figure}[h!]
    \centering
    \includegraphics[width=\textwidth]{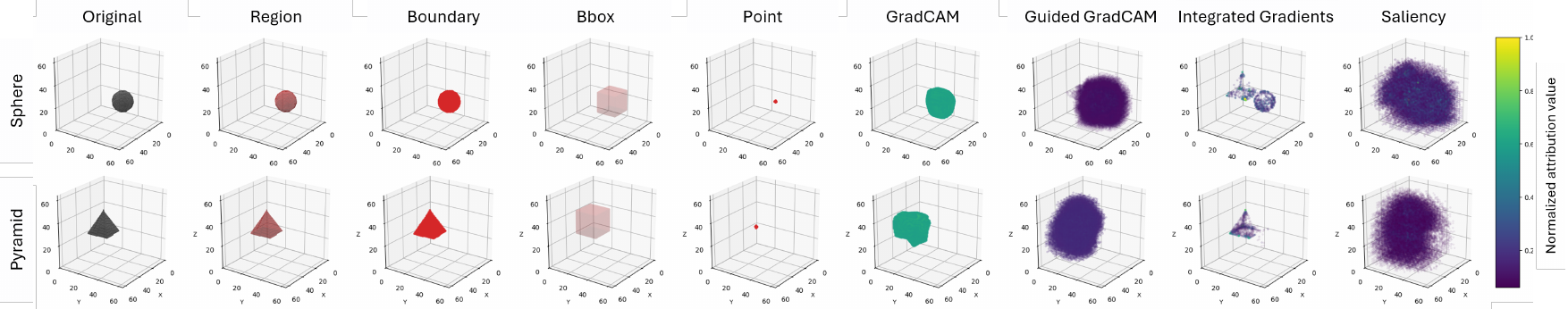}
    \caption{Synthetic 3D examples showing multiple ground-truth (GT) representations (region, boundary, bounding box, and point) derived from the same objects (sphere and pyramid), together with attribution maps from representative XAI methods.}
    \label{fig:xai-gt_misalignment}
\end{figure}

 As illustrated in Figure~\ref{fig:xai-gt_misalignment}, different GT formulations derived from the same object emphasize distinct spatial characteristics, leading to varying alignment outcomes when compared with explanations exhibiting diverse spatial behaviors. For instance, region-like explanations tend to align better with mask-based GTs, while more diffuse or edge-focused explanations may correspond differently depending on the chosen representation. This demonstrates that GT selection is not a neutral choice, but a key component of the evaluation configuration that directly influences alignment scores and their interpretation. To analyze the effect of GT in a controlled manner, we construct multiple GT representations from the same underlying target.

\subsection{Alignment Metrics}

To evaluate alignment between explanations and ground truth, we adopt a taxonomy-driven selection of complementary metrics, where each metric captures a distinct aspect of spatial agreement. Specifically, we consider five metric groups: \emph{overlap-based}, \emph{point-based}, \emph{distance-based}, \emph{mass-based}, and \emph{leakage-based}~\citep{maier2024metrics, nauta2023anecdotal}. These respectively quantify spatial overlap, peak localization, structural proximity, attribution concentration, and leakage outside the ground truth.

Following this taxonomy, we select representative metrics from each group: Intersection over Union (IoU), Pointing Game (PG), Energy-based Pointing Game (EPG), Background Energy Ratio (BER), and Hausdorff Distance (HD), respectively.

\paragraph{Soft vs.\ Hard Formulations.}
Metrics are computed either on binarized supports (hard) or continuous attribution maps (soft). Hard formulations operate on thresholded explanations $\tilde{A}$ and evaluate spatial localization (\emph{where} the explanation lies), while soft formulations operate directly on $A$ and capture attribution distribution (\emph{how strongly} relevance is expressed). The distinction between hard and soft formulations is critical, as thresholding introduces sensitivity to hyperparameters and may discard diffuse but meaningful attribution signals.

To ensure consistency across formulations, we include several adapted variants (marked with $^*$ in Table~\ref{tab:alignment_metrics_summary}). First, we introduce a soft variant of the Pointing Game using an exponent $\alpha$ to emphasize high-attribution peaks while retaining a continuous formulation. Second, we define Background Energy Ratio (BER) to complement EPG, explicitly capturing attribution leakage outside the ground truth. Finally, we adopt a directed soft formulation of Hausdorff Distance (HD) inspired by~\citep{huttenlocher2002comparing}, which measures the one-way directed attribution-weighted distance from ground truth to explanation. Unlike symmetric hard HD, our proposed soft formulation measures how well attribution mass covers the GT structure, enabling continuous distance-based evaluation without thresholding.

\begin{table*}[t]
\centering
\caption{Summary of alignment metrics: Intersection over Union (IoU), Pointing Game (PG), Energy-based Pointing Game (EPG), Background Energy Ratio (BER), and Hausdorff Distance (HD) with corresponding hard (threshold-based) and soft (continuous) formulations. $A$ denotes the attribution map, $\tilde{A}$ the binarized and thresholded explanation support, and $G$ the ground-truth region. $^*$ denotes derived or adapted formulations introduced in this work for consistency across soft and hard evaluation settings.}
\label{tab:alignment_metrics_summary}
\vspace{0.5em}
\scriptsize
\setlength{\tabcolsep}{4pt}
\renewcommand{\arraystretch}{1.1}
\begin{tabularx}{\textwidth}{
L{1.0cm}
L{0.9cm}
C{3.1cm}
>{\raggedright\arraybackslash}X
L{1.5cm}
C{0.9cm}
C{1.0cm}
}
\toprule
\textbf{Metric} & \textbf{Variant} & \textbf{Formula} & \textbf{Question answered} & \textbf{Measures} & \textbf{Range} & \textbf{Preferred} \\
\midrule

\multirow{2}{*}[-3.0ex]{IoU \cite{rezatofighi2019generalized}}
& Hard
& $\displaystyle \frac{\sum_i \tilde{A}_i G_i}{\sum_i (\tilde{A}_i + G_i - \tilde{A}_i G_i)}$
& How much do the explanation support and ground truth overlap?
& Overlap
& $[0,1]$
& Higher \\ \cmidrule {2-7} \addlinespace[1.0ex]
& Soft
& $\displaystyle \frac{\sum_i \min(A_i,G_i)}{\sum_i \max(A_i,G_i)}$
& How much do the attribution map and ground truth overlap proportionally?
& Soft overlap
& $[0,1]$
& Higher \\
\midrule

\multirow{2}{*}[-2.4ex]{PG \cite{zhang2018top}}
& Hard
& $\displaystyle \mathbb{1}[\arg\max_i A_i \in G]$
& Is the most important point inside the ground truth?
& Peak Correctness
& $\{0,1\}$
& Higher \\ \cmidrule {2-7}
& Soft*
& $\displaystyle \frac{\sum_i A_i^\alpha G_i}{\sum_i A_i^\alpha}$
& Do high-attribution peaks lie inside the ground truth?
& Peak Alignment
& $[0,1]$
& Higher \\
\midrule

\multirow{2}{*}[-2.4ex]{EPG \cite{wang2020score}}
& Hard
& $\displaystyle \frac{\sum_i \tilde{A}_i G_i}{\sum_i \tilde{A}_i}$
& How much of the explanation support lies inside the ground truth?
& Precision
& $[0,1]$
& Higher \\ \cmidrule {2-7} \addlinespace[1.0ex]
& Soft
& $\displaystyle \frac{\sum_i A_i G_i}{\sum_i A_i}$
& How much attribution mass lies inside the ground truth?
& Mass in GT
& $[0,1]$
& Higher \\
\midrule

\multirow{2}{*}[-2.6ex]{BER*}
& Hard
& $\displaystyle 1-\frac{\sum_i \tilde{A}_i G_i}{\sum_i \tilde{A}_i}$
& How much of the explanation support lies outside the ground truth?
& Support Leakage
& $[0,1]$
& Lower \\ \cmidrule {2-7} \addlinespace[1.0ex]
& Soft
& $\displaystyle 1-\frac{\sum_i A_i G_i}{\sum_i A_i}$
& How much attribution mass lies outside the ground truth?
& Mass Leakage
& $[0,1]$
& Lower \\
\midrule

\multirow{2}{*}[-2.5ex]{HD \cite{huttenlocher2002comparing}}
& Hard
& $\displaystyle \max\!\big(d(\tilde{A},G),\, d(G,\tilde{A})\big)$
& How large is the structural mismatch between explanation and ground truth?
& Structural Distance
& $[0,\infty)$
& Lower \\ \cmidrule {2-7}
& Soft*
& $\displaystyle \frac{1}{|G|}\sum_{g\in G}\sum_i A_i\, d(g,i)$
& How far is the ground truth from the explanation on average?
& Directed Distance
& $[0,\infty)$
& Lower \\
\bottomrule
\end{tabularx}
\end{table*}
\vspace{-4pt}

Table~\ref{tab:alignment_metrics_summary} summarizes all metrics, their formulations, and the corresponding evaluation objectives. Together, these metrics provide a comprehensive and complementary view of alignment under different assumptions about explanation behavior and ground-truth structure.

\section{Experimental Results}
In this section, we empirically analyze how alignment evaluation outcomes vary across different evaluation configurations. Building on the configuration space defined in Section~\ref{sec:methodology}, we systematically investigate the interaction between XAI methods, GT representations, and alignment metrics. Our goal is twofold: (i) to identify consistent alignment patterns across datasets and (ii) to quantify the extent to which evaluation outcomes depend on configuration choices. We further analyze the stability of these outcomes under different metric formulations and threshold settings.

\subsection{Experimental Setup}

We evaluate our framework on four datasets covering controlled and real-world datasets: 2D synthetic, 3D synthetic, BRISC2025~\citep{fateh2026brisc}, and MosMed~\citep{morozov2020mosmeddata}. Synthetic datasets enable full control over object structure and GT representations, while real datasets provide realistic variability in medical imaging scenarios. The 2D synthetic dataset consists of multi-label images composed of simple geometric shapes (e.g., circle, square, triangle), where each sample may contain multiple objects with known spatial properties. From these objects, we derive multiple GT representations (region, boundary, bounding box, and point). The 3D synthetic dataset extends these representations to volumetric data (e.g., sphere, cube, pyramid) with voxel-based representations, enabling evaluation in a fully controlled 3D setting. All tasks are treated as classification, with explanation analysis performed at the instance level. (Additional details regarding datasets, train/test splits, and predictive model configurations are provided in Appendix~\ref{appx:model_train_dataset}.)

We use attribution-based XAI methods implemented in Captum~\citep{kokhlikyan2020captum} with default configurations to ensure a consistent comparison across methods; method-specific settings and attribution post-processing details are provided in the public repository. We evaluate alignment using five metric families (IoU, PG, EPG, BER, HD), each in both soft and hard formulations as defined in Table~\ref{tab:alignment_metrics_summary}.

We use standard convolutional architectures suitable for attribution-based explanations. For synthetic datasets, we train lightweight CNN classifiers consisting of four convolutional blocks with pooling, using 2D and 3D variants for image and volumetric data, respectively. For BRISC2025, we fine-tune a pretrained DenseNet121 of PyTorch~\citep{paszke2019pytorch}. For MosMed, we use a pretrained 3D DenseNet from MONAI~\citep{cardoso2022monai} with transfer learning. All models achieve reliable predictive performance (macro-F1 above $0.7$ across datasets), ensuring that explanations are generated from sufficiently accurate models. 

\paragraph{Ground-Truth Construction.}
To enable controlled comparison across evaluation configurations, we derive multiple GT representations (i.e., region, boundary, bounding box, and point) from a common region mask. Region GT corresponds to the cleaned binary mask. Boundary GT is obtained via morphological operations (dilation–erosion difference). Bounding boxes are generated per connected component using min–max spatial extents and rasterized into masks. In 2D, Point GT is centered at a distance-transform maximum and rasterized as a filled radius-two disk. In 3D, it is centered at the rounded component centroid and rasterized as a seven-voxel, 6-connected axial marker. All representations were generated consistently from the same underlying regions to isolate the effect of GT formulation on alignment outcomes.

\subsection{XAI Output Behaviors}

We analyze the spatial behavior of 15 attribution-based XAI methods including gradient-based approaches (Saliency~\citep{simonyan2013deep}, Integrated Gradients~\citep{sundararajan2017axiomatic}, Input$\times$Gradient, SmoothGrad~\citep{smilkov2017smoothgrad}), CAM-based methods (Grad-CAM~\citep{selvaraju2017grad}, Grad-CAM++~\citep{chattopadhay2018gradcamplusplus}, Score-CAM~\citep{wang2020scorecam}, Eigen-CAM~\citep{muhammad2020eigen}, Layer-CAM~\citep{jiang2021layercam}), relevance-based methods (LRP~\citep{bach2015pixel}), perturbation-based methods (Occlusion and Feature Ablation~\citep{zeiler2014visualizing}), DeepLIFT~\citep{shrikumar2017learning}, and hybrid approaches such as Guided Backpropagation and Guided Grad-CAM~\citep{springenberg2015striving, selvaraju2017grad}.

\begin{figure}[h!]
    \centering
    \includegraphics[width=\linewidth]{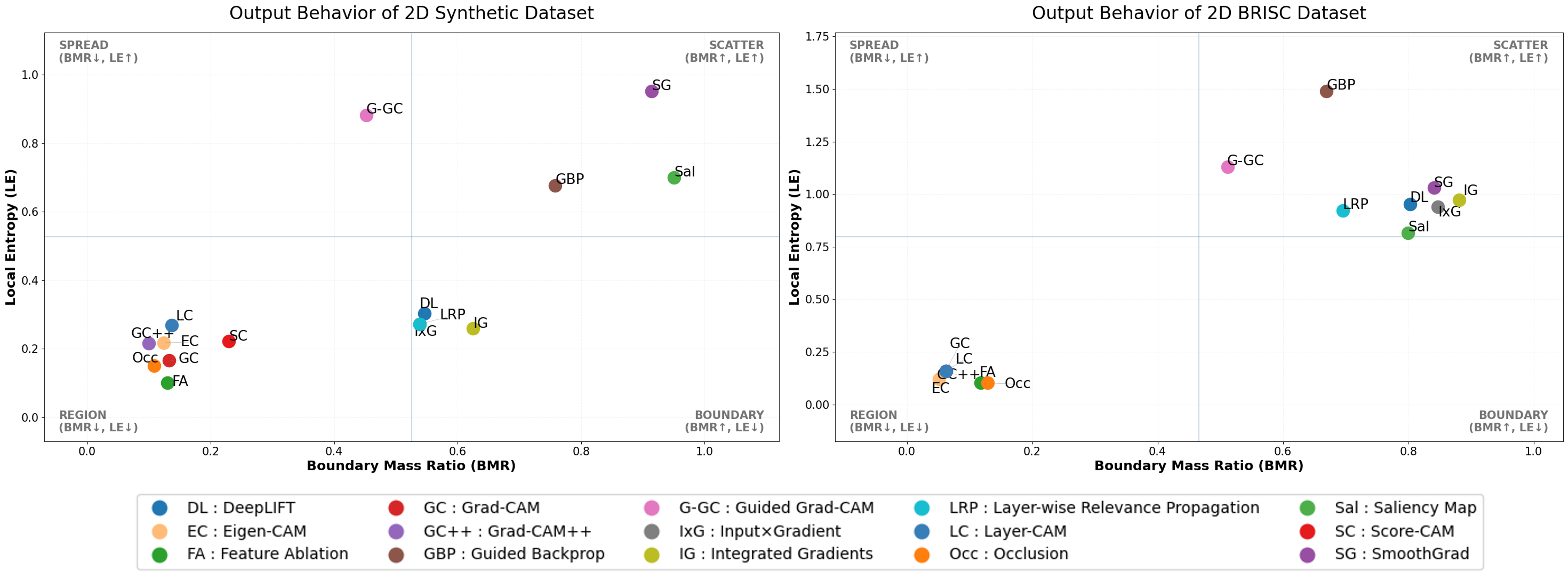}
    \caption{
    Output behavior of XAI methods for the 2D synthetic dataset (left) and the BRISC2025 dataset (right), characterized by local entropy (LE) and boundary mass ratio (BMR).
    }
    \label{fig:xai_behavior}
\end{figure}

Figure~\ref{fig:xai_behavior} illustrates their behavior in the BMR–LE space for a controlled 2D synthetic dataset and the BRISC2025 dataset. Within each dataset, the methods occupy different regions of the continuous BMR--LE descriptor space, indicating different spatial attribution behaviors under the evaluated conditions.

Across datasets, we observe consistent method-specific behaviors. CAM-based methods (e.g., Grad-CAM and its variants) consistently exhibit region-focused patterns, reflecting their coarse spatial localization. In contrast, gradient-based methods display more diverse behaviors: Guided Grad-CAM spans spread and scatter regions due to its combination of localization and high-frequency gradients, while Integrated Gradients produces more distributed attributions, placing it in boundary or scatter quadrants depending on the dataset. Importantly, some methods do not correspond to a single behavior type but shift across regions depending on data and problem characteristics (i.e., multi-label classification), indicating that XAI behavior is context-dependent.

Based on this analysis, we select four representative methods, i.e., Grad-CAM, Guided Grad-CAM, Integrated Gradients, and Saliency, to ensure that subsequent alignment evaluation captures a diverse spectrum of attribution behaviors.

\subsection{Alignment Outcomes Across Configurations}

We analyze how alignment outcomes vary across XAI methods, GT representations, metric families, and formulations across all datasets. Since alignment metrics differ in scale and interpretation (e.g., bounded overlap vs. unbounded distances), raw scores are not directly comparable. We therefore standardize scores using z-score normalization within each metric and focus on ranking-based comparisons. This transforms the analysis from absolute metric comparison to relative evaluation preference analysis across configurations. Since BER is the exact complement of EPG, the ranking and flow analyses use the four non-redundant metrics IoU, PG, EPG, and HD.

\begin{figure*}[h!]
    \centering
    \includegraphics[width=\linewidth]{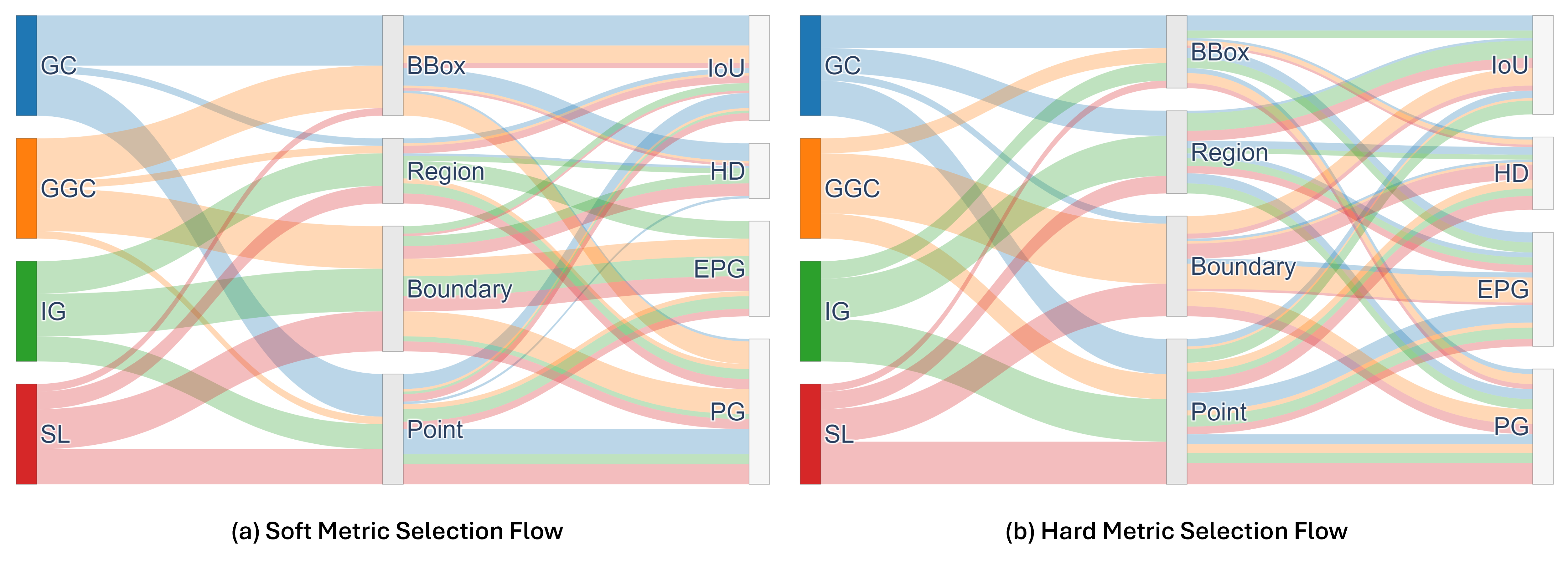}
    \caption{
    Alignment selection flows across XAI methods, ground-truth (GT) representations, and metric families under soft (a) and hard (b) formulations. Edge thickness indicates cumulative rank-weighted selection frequency across datasets.
    }
    \label{fig:sankey_comparison}
\end{figure*}

For each dataset and formulation, we rank XAI–GT pairs based on mean standardized scores and select the top-2 GT representations per XAI method. These are further expanded with their top supporting metrics to form XAI–GT–metric triples. The resulting positions receive descending rank weights, which are accumulated across datasets to construct the soft and hard selection flows. Figure~\ref{fig:sankey_comparison} presents aggregated selection flows under soft and hard formulations. The flow-based formulation enables analysis of interaction structure across evaluation components rather than isolated pairwise comparisons. This results in configuration-dependent alignment patterns between hard and soft metric families rather than consistent metric ranking among them.

It can be noticed from Figure~\ref{fig:sankey_comparison} that the soft
formulation preserves three prominent compatibility patterns: Bounding
Box--IoU, Point--PG, and Boundary--EPG. At the behavioral level, the
compact, low-LE attribution pattern exhibited by Grad-CAM contributes
strongly to the Bounding Box--IoU pathway. Methods occupying higher-LE
or more boundary-concentrated parts of the descriptor space divide
between Point--PG and Boundary--EPG rather than sharing one preferred
configuration. Thus, attribution behavior informs the evaluation
pathway but does not determine a unique GT or metric.

Under the hard formulation, the flows become more concentrated around
Point--PG. Integrated Gradients and Saliency shift more strongly toward
Point, indicating that thresholding emphasizes their retained
high-attribution locations, whereas Guided Grad-CAM retains a strong
Boundary pathway and Grad-CAM remains divided between Point and Bounding
Box. Changing the formulation therefore changes which aspect of an
attribution behavior dominates the alignment evaluation.

\paragraph{Configuration Dependency.}
We quantify differences between soft and hard formulations using Kendall’s $\tau$, mean rank shift, top-1 change rate, and top-2 overlap. The 2D synthetic dataset shows the highest agreement ($\tau=0.67$), while BRISC2025 and the 3D synthetic dataset show moderate agreement ($\tau=0.42$ for both). MosMed exhibits the strongest formulation dependence, with negative ranking agreement ($\tau=-0.17$) and the largest mean rank shift ($1.38$).

Critically, the leading GT representation changes for $25\%$ of the XAI methods in
the 2D synthetic and BRISC2025 datasets, $50\%$ in the 3D synthetic dataset, and $75\%$ in MosMed. Top-2 overlap ranges from $0.50$ to $0.88$, indicating that the set of plausible GT representations is generally more stable than their ordering. Nevertheless, the changes in both the leading selection and the complete rankings confirm that the preferred evaluation configuration depends on whether a soft or hard metric formulation is used (see Appendix~\ref{appx:ranking_agreement})

These differences in rankings between soft and hard metrics correspond to structural shifts in the flow as well (See Figure~\ref{fig:sankey_comparison}). Soft metrics preserve attribution distributions and place greater emphasis on coverage- and mass-based alignment (IoU, EPG), while hard metrics emphasize peak
localization after thresholding, increasing the relative importance of point-based evaluation (PG). Thus, metric formulation does not merely rescale scores, but changes the aspect of the explanation being evaluated.

\paragraph{Statistical Validation.}
We assess robustness of alignment rankings via bootstrap resampling ($B=1000$). Results remain consistent across resamples, confirming that observed patterns are not driven by sample variability (see Appendix~\ref{appx:bootstrap}).

\subsection{Threshold Sensitivity}

We analyze effect of thresholding on hard metrics by evaluating top-$k$ thresholds $\mathcal{T} = \{0.01, 0.05, 0.10, 0.15, 0.20, 0.30, 0.50, 0.75, 0.90, 1.00\}$. For each configuration, we compute the mean metric value across samples for each threshold, resulting in a response curve that captures how alignment behavior changes with respect to thresholding. Sensitivity is measured via interquartile range (IQR), and stability via the proportion of thresholds exhibiting low variation.

We observe systematic differences across metric families. Point-based metrics (PG) are highly stable and insensitive to thresholding, as they depend on peak locations. Overlap- and mass-based metrics (IoU, EPG, BER) show moderate sensitivity, reflecting dependence on spatial support. Distance-based metrics (HD) exhibit the highest sensitivity, where small threshold changes significantly affect structural relationships. Sensitivity further depends on XAI behavior and GT choice. Region-based explanations are more stable under overlap metrics, while diffuse explanations are highly sensitive due to loss of low-intensity attribution. Similarly, point-based GTs are more stable, whereas region and boundary GTs introduce higher variability. Detailed sensitivity curves and stability analyses for all datasets are provided in the Appendix~\ref{appx:sensitivity},\ref{appx:hard-sensitivity-stability}.

Across datasets, synthetic data shows greater stability, while real-world and 3D data exhibit stronger threshold dependence due to increased structural complexity. These results demonstrate that hard metrics are not threshold-invariant, and threshold sensitivity must explicitly be analyzed when evaluating alignment.


\section{Discussion and Conclusion}
This paper studies alignment evaluation as a structured measurement process composed of interacting evaluation assumptions, including GT representation, metric family, evaluation formulation, thresholding, and attribution structure. Through flow-based analysis across XAI methods, GT representations, and metric families, we identify systematic and interpretable alignment patterns that reveal how different evaluation configurations emphasize different properties of explanation behavior. These findings demonstrate that alignment evaluation should not be interpreted as a single universal measure of explanation quality, but rather as a structured evaluation process whose conclusions depend on how alignment is defined and measured based on the configuration component characteristics. In this study, alignment therefore denotes specifically the spatial correspondence between an attribution map and a selected GT representation under a stated metric and formulation. It can provide a scalable proxy for coherence with expert-provided annotations, but a high alignment score does not establish explanation faithfulness, predictive correctness, usefulness, or overall human-centered quality; those properties require complementary model-centered and human-centered evaluation.

Our results reveal that region-focused methods align with region-based GTs under overlap metrics, boundary-focused methods align with boundary GTs under point- and mass-based metrics, and diffuse methods align with point-based representations. These relationships recur across both real-world and synthetic datasets used in experiments, although their strength and the behavioral position of individual XAI methods vary with
the dataset, metric, and formulation. Additionally, it is revealed by the results that soft metrics capture attribution distribution and favor coverage-based alignment, whereas hard metrics emphasize peak responses due to thresholding, particularly for complex real-world and 3D datasets.

Extending the framework to additional modalities, tasks, and XAI methods represents an important direction for future research.

\paragraph{Alignment Evaluation Guidelines.}
Based on the observed configuration-dependent behavior, we derive the following recommendations for alignment evaluation of XAI methods:

\begin{itemize}
    \item \textbf{Match metrics to explanation behavior:} Region-focused explanations are better captured by overlap-based metrics and region GTs, whereas boundary-focused or diffuse explanations require point- or mass-based evaluation.
    
    \item \textbf{Use complementary evaluation formulations:} Soft metrics capture attribution distribution, while hard metrics emphasize localization behavior. Similarly, different metric families capture complementary structural properties and may produce conflicting rankings. Relying on a single formulation or metric family may therefore lead to incomplete evaluation conclusions.
    
    \item \textbf{Report threshold stability:} Hard metrics are inherently threshold-sensitive; stability analysis should therefore accompany threshold-based evaluation.
\end{itemize}

Future work will extend the proposed alignment evaluation framework to additional modalities, tasks, and XAI methods to further investigate alignment patterns across broader evaluation settings.

\paragraph{Code and data availability.}
The implementation used in this study, including synthetic-data generation,
ground-truth construction, alignment metrics, evaluation analyses, and figure
generation, is publicly available at
\url{https://github.com/gizemkaragozz/Unpacking-the-Evaluators}.
The medical datasets are accessed from their original public sources under
their respective terms of use.

\section*{Acknowledgments}

This work was conducted in the context of the ITEA 3 Secure e-Health project (project number 20050), ``Privacy-Preserving Cross-Organizational Data Analysis in the Healthcare Sector.''

The authors declare no competing interests.

{
\small
\medskip

\bibliographystyle{plain}
\bibliography{references}

@article{miller2019explanation,
  title={Explanation in artificial intelligence: Insights from the social sciences},
  author={Miller, Tim},
  journal={Artificial intelligence},
  volume={267},
  pages={1--38},
  year={2019},
  publisher={Elsevier}
}

@article{rong2023towards,
  title={Towards human-centered explainable ai: A survey of user studies for model explanations},
  author={Rong, Yao and Leemann, Tobias and Nguyen, Thai-Trang and Fiedler, Lisa and Qian, Peizhu and Unhelkar, Vaibhav and Seidel, Tina and Kasneci, Gjergji and Kasneci, Enkelejda},
  journal={IEEE transactions on pattern analysis and machine intelligence},
  volume={46},
  number={4},
  pages={2104--2122},
  year={2023},
  publisher={IEEE}
}

@inproceedings{chase2024attribution,
  title={Attribution quality metrics with magnitude alignment},
  author={Chase Walker, Dominic Simon and Chen, Kenny and Ewetz, Rickard},
  booktitle={Proceedings of the Thirty-Third International Joint Conference on Artificial Intelligence, IJCAI-24},
  pages={530--538},
  year={2024}
}

@article{arrieta2020explainable,
  title={Explainable Artificial Intelligence (XAI): Concepts, taxonomies, opportunities and challenges toward responsible AI},
  author={Arrieta, Alejandro Barredo and D{\'\i}az-Rodr{\'\i}guez, Natalia and Del Ser, Javier and Bennetot, Adrien and Tabik, Siham and Barbado, Alberto and Garc{\'\i}a, Salvador and Gil-L{\'o}pez, Sergio and Molina, Daniel and Benjamins, Richard and others},
  journal={Information fusion},
  volume={58},
  pages={82--115},
  year={2020},
  publisher={Elsevier}
}

@article{deng2026attribution,
  title={Attribution explanations for deep neural networks: A theoretical perspective},
  author={Deng, Huiqi and Pei, Hongbin and Zhang, Quanshi and Du, Mengnan},
  journal={IEEE Transactions on Pattern Analysis and Machine Intelligence},
  year={2026},
  publisher={IEEE}
}

@inproceedings{speith2022review,
  title={A review of taxonomies of explainable artificial intelligence (XAI) methods},
  author={Speith, Timo},
  booktitle={Proceedings of the 2022 ACM conference on fairness, accountability, and transparency},
  pages={2239--2250},
  year={2022}
}

@article{hedstrom2023quantus,
  title={Quantus: An explainable ai toolkit for responsible evaluation of neural network explanations and beyond},
  author={Hedstr{\"o}m, Anna and Weber, Leander and Krakowczyk, Daniel and Bareeva, Dilyara and Motzkus, Franz and Samek, Wojciech and Lapuschkin, Sebastian and H{\"o}hne, Marina M-C},
  journal={Journal of Machine Learning Research},
  volume={24},
  number={34},
  pages={1--11},
  year={2023}
}

@inproceedings{mullen2019comparing,
  title={Comparing the effects of annotation type on machine learning detection performance},
  author={Mullen Jr, James F and Tanner, Franklin R and Sallee, Phil A},
  booktitle={Proceedings of the ieee/cvf conference on computer vision and pattern recognition workshops},
  pages={0--0},
  year={2019}
}

@article{agnew2023quantifying,
  title={Quantifying the effects of ground truth annotation quality on object detection and instance segmentation performance},
  author={Agnew, Cathaoir and Eising, Ciar{\'a}n and Denny, Patrick and Scanlan, Anthony and Van De Ven, Pepijn and Grua, Eoin M},
  journal={IEEE Access},
  volume={11},
  pages={25174--25188},
  year={2023},
  publisher={IEEE}
}

@article{arras2022clevr,
  title={CLEVR-XAI: A benchmark dataset for the ground truth evaluation of neural network explanations},
  author={Arras, Leila and Osman, Ahmed and Samek, Wojciech},
  journal={Information Fusion},
  volume={81},
  pages={14--40},
  year={2022},
  publisher={Elsevier}
}

@inproceedings{le2023benchmarking,
  title={Benchmarking eXplainable AI-A Survey on Available Toolkits and Open Challenges.},
  author={Le, Phuong Quynh and Nauta, Meike and Van Bach Nguyen, Shreyasi Pathak and Pathak, Shreyasi and Schl{\"o}tterer, J{\"o}rg and Seifert, Christin},
  booktitle={IJCAI},
  pages={6665--6673},
  year={2023}
}

@inproceedings{sundararajan2017axiomatic,
  title={Axiomatic attribution for deep networks},
  author={Sundararajan, Mukund and Taly, Ankur and Yan, Qiqi},
  booktitle={International conference on machine learning},
  pages={3319--3328},
  year={2017},
  organization={PMLR}
}

@inproceedings{cheng2021boundary,
  title={Boundary IoU: Improving object-centric image segmentation evaluation},
  author={Cheng, Bowen and Girshick, Ross and Doll{\'a}r, Piotr and Berg, Alexander C and Kirillov, Alexander},
  booktitle={Proceedings of the IEEE/CVF conference on computer vision and pattern recognition},
  pages={15334--15342},
  year={2021}
}

@article{klein2024navigating,
  title={Navigating the maze of explainable ai: A systematic approach to evaluating methods and metrics},
  author={Klein, Lukas and L{\"u}th, Carsten and Schlegel, Udo and Bungert, Till and El-Assady, Mennatallah and J{\"a}ger, Paul},
  journal={Advances in Neural Information Processing Systems},
  volume={37},
  pages={67106--67146},
  year={2024}
}

@inproceedings{selvaraju2017grad,
  title={Grad-cam: Visual explanations from deep networks via gradient-based localization},
  author={Selvaraju, Ramprasaath R and Cogswell, Michael and Das, Abhishek and Vedantam, Ramakrishna and Parikh, Devi and Batra, Dhruv},
  booktitle={Proceedings of the IEEE international conference on computer vision},
  pages={618--626},
  year={2017}
}

@article{smilkov2017smoothgrad,
  title={Smoothgrad: removing noise by adding noise},
  author={Smilkov, Daniel and Thorat, Nikhil and Kim, Been and Vi{\'e}gas, Fernanda and Wattenberg, Martin},
  journal={arXiv preprint arXiv:1706.03825},
  year={2017}
}

@article{adebayo2018sanity,
  title={Sanity checks for saliency maps},
  author={Adebayo, Julius and Gilmer, Justin and Muelly, Michael and Goodfellow, Ian and Hardt, Moritz and Kim, Been},
  journal={Advances in neural information processing systems},
  volume={31},
  year={2018}
}

@article{hooker2019benchmark,
  title={A benchmark for interpretability methods in deep neural networks},
  author={Hooker, Sara and Erhan, Dumitru and Kindermans, Pieter-Jan and Kim, Been},
  journal={Advances in neural information processing systems},
  volume={32},
  year={2019}
}

@inproceedings{petsiuk2018rise,
  title={RISE: Randomized Input Sampling for Explanation of Black-box Models},
  author={Petsiuk, Vitali and Das, Abir and Saenko, Kate},
  booktitle={BMVC},
  year={2018}
}

@inproceedings{chattopadhay2018gradcamplusplus,
  title={Grad-cam++: Generalized gradient-based visual explanations for deep convolutional networks},
  author={Chattopadhay, Aditya and Sarkar, Anirban and Howlader, Prantik and Balasubramanian, Vineeth N},
  booktitle={2018 IEEE winter conference on applications of computer vision (WACV)},
  pages={839--847},
  year={2018},
  organization={IEEE}
}

@inproceedings{deyoung2020eraser,
  title={ERASER: A benchmark to evaluate rationalized NLP models},
  author={DeYoung, Jay and Jain, Sarthak and Rajani, Nazneen Fatema and Lehman, Eric and Xiong, Caiming and Socher, Richard and Wallace, Byron C},
  booktitle={Proceedings of the 58th annual meeting of the association for computational linguistics},
  pages={4443--4458},
  year={2020}
}

@article{doshi2017towards,
  title={Towards a rigorous science of interpretable machine learning},
  author={Doshi-Velez, Finale and Kim, Been},
  journal={arXiv preprint arXiv:1702.08608},
  year={2017}
}

@article{nauta2023anecdotal,
  title={From anecdotal evidence to quantitative evaluation methods: A systematic review on evaluating explainable ai},
  author={Nauta, Meike and Trienes, Jan and Pathak, Shreyasi and Nguyen, Elisa and Peters, Michelle and Schmitt, Yasmin and Schl{\"o}tterer, J{\"o}rg and Van Keulen, Maurice and Seifert, Christin},
  journal={ACM Computing Surveys},
  volume={55},
  number={13s},
  pages={1--42},
  year={2023},
  publisher={ACM New York, NY}
}

@inproceedings{tonekaboni2019clinicians,
  title={What clinicians want: contextualizing explainable machine learning for clinical end use},
  author={Tonekaboni, Sana and Joshi, Shalmali and McCradden, Melissa D and Goldenberg, Anna},
  booktitle={Machine learning for healthcare conference},
  pages={359--380},
  year={2019},
  organization={PMLR}
}

@article{amann2020explainability,
  title={Explainability for artificial intelligence in healthcare: a multidisciplinary perspective},
  author={Amann, Julia and Blasimme, Alessandro and Vayena, Effy and Frey, Dietmar and Madai, Vince I and Precise4Q Consortium},
  journal={BMC medical informatics and decision making},
  volume={20},
  number={1},
  pages={310},
  year={2020},
  publisher={Springer}
}

@article{ghassemi2021false,
  title={The false hope of current approaches to explainable artificial intelligence in health care},
  author={Ghassemi, Marzyeh and Oakden-Rayner, Luke and Beam, Andrew L},
  journal={The lancet digital health},
  volume={3},
  number={11},
  pages={e745--e750},
  year={2021},
  publisher={Elsevier}
}

@inproceedings{rezatofighi2019generalized,
  title={Generalized intersection over union: A metric and a loss for bounding box regression},
  author={Rezatofighi, Hamid and Tsoi, Nathan and Gwak, JunYoung and Sadeghian, Amir and Reid, Ian and Savarese, Silvio},
  booktitle={Proceedings of the IEEE/CVF conference on computer vision and pattern recognition},
  pages={658--666},
  year={2019}
}

@article{zhang2018top,
  title={Top-down neural attention by excitation backprop},
  author={Zhang, Jianming and Bargal, Sarah Adel and Lin, Zhe and Brandt, Jonathan and Shen, Xiaohui and Sclaroff, Stan},
  journal={International Journal of Computer Vision},
  volume={126},
  number={10},
  pages={1084--1102},
  year={2018},
  publisher={Springer}
}

@inproceedings{wang2020score,
  title={Score-CAM: Score-weighted visual explanations for convolutional neural networks},
  author={Wang, Haofan and Wang, Zifan and Du, Mengnan and Yang, Fan and Zhang, Zijian and Ding, Sirui and Mardziel, Piotr and Hu, Xia},
  booktitle={Proceedings of the IEEE/CVF conference on computer vision and pattern recognition workshops},
  pages={24--25},
  year={2020}
}

@article{huttenlocher2002comparing,
  title={Comparing images using the Hausdorff distance},
  author={Huttenlocher, Daniel P and Klanderman, Gregory A. and Rucklidge, William J},
  journal={IEEE Transactions on pattern analysis and machine intelligence},
  volume={15},
  number={9},
  pages={850--863},
  year={2002},
  publisher={IEEE}
}

@article{fateh2026brisc,
  title={BRISC: Annotated dataset for brain tumor segmentation and classification},
  author={Fateh, Amirreza and Rezvani, Yasin and Moayedi, Sara and Rezvani, Sadjad and Fateh, Fatemeh and Fateh, Mansoor and Abolghasemi, Vahid},
  journal={Scientific Data},
  year={2026},
  publisher={Nature Publishing Group UK London}
}

@article{morozov2020mosmeddata,
  title={Mosmeddata: Chest ct scans with covid-19 related findings dataset},
  author={Morozov, Sergey P and Andreychenko, Anna E and Pavlov, Nikolay A and Vladzymyrskyy, AV and Ledikhova, Natalya V and Gombolevskiy, Victor A and Blokhin, Ivan A and Gelezhe, Pavel B and Gonchar, AV and Chernina, V Yu},
  journal={arXiv preprint arXiv:2005.06465},
  year={2020}
}

@article{kokhlikyan2020captum,
  title={Captum: A unified and generic model interpretability library for pytorch},
  author={Kokhlikyan, Narine and Miglani, Vivek and Martin, Miguel and Wang, Edward and Alsallakh, Bilal and Reynolds, Jonathan and Melnikov, Alexander and Kliushkina, Natalia and Araya, Carlos and Yan, Siqi and others},
  journal={arXiv preprint arXiv:2009.07896},
  year={2020}
}

@article{wu2013local,
  title={Local Shannon entropy measure with statistical tests for image randomness},
  author={Wu, Yue and Zhou, Yicong and Saveriades, George and Agaian, Sos and Noonan, Joseph P and Natarajan, Premkumar},
  journal={Information Sciences},
  volume={222},
  pages={323--342},
  year={2013},
  publisher={Elsevier}
}

@article{maier2024metrics,
  title={Metrics reloaded: recommendations for image analysis validation},
  author={Maier-Hein, Lena and Reinke, Annika and Godau, Patrick and Tizabi, Minu D and Buettner, Florian and Christodoulou, Evangelia and Glocker, Ben and Isensee, Fabian and Kleesiek, Jens and Kozubek, Michal and others},
  journal={Nature methods},
  volume={21},
  number={2},
  pages={195--212},
  year={2024},
  publisher={Nature Publishing Group US New York}
}

@article{cardoso2022monai,
  title={Monai: An open-source framework for deep learning in healthcare},
  author={Cardoso, M Jorge and Li, Wenqi and Brown, Richard and Ma, Nic and Kerfoot, Eric and Wang, Yiheng and Murrey, Benjamin and Myronenko, Andriy and Zhao, Can and Yang, Dong and others},
  journal={arXiv preprint arXiv:2211.02701},
  year={2022}
}

@article{paszke2019pytorch,
  title={Pytorch: An imperative style, high-performance deep learning library},
  author={Paszke, Adam and Gross, Sam and Massa, Francisco and Lerer, Adam and Bradbury, James and Chanan, Gregory and Killeen, Trevor and Lin, Zeming and Gimelshein, Natalia and Antiga, Luca and others},
  journal={Advances in neural information processing systems},
  volume={32},
  year={2019}
}

@inproceedings{simonyan2013deep,
  title={Deep Inside Convolutional Networks: Visualising Image Classification Models and Saliency Maps},
  author={Simonyan, Karen and Vedaldi, Andrea and Zisserman, Andrew},
  booktitle={ICLR Workshop},
  year={2014}
}

@inproceedings{wang2020scorecam,
  title={Score-CAM: Score-Weighted Visual Explanations for Convolutional Neural Networks},
  author={Wang, Haofan and Wang, Zifan and Du, Mengnan and Yang, Fan and Zhang, Zijian and Ding, Siyu and Mardziel, Piotr and Hu, Xia},
  booktitle={CVPR Workshops},
  year={2020}
}

@inproceedings{muhammad2020eigen,
  title={Eigen-CAM: Class Activation Map using Principal Components},
  author={Muhammad, Muhammad Bilal and Yeasin, Mohammed},
  booktitle={ICPR},
  year={2020}
}

@inproceedings{jiang2021layercam,
  title={LayerCAM: Exploring Hierarchical Class Activation Maps for Localization},
  author={Jiang, Peng-Tao and Zhang, Chang-Bin and Hou, Qibin and Cheng, Ming-Ming and Wei, Yunchao},
  booktitle={IEEE TIP},
  year={2021}
}

@inproceedings{bach2015pixel,
  title={On Pixel-Wise Explanations for Non-Linear Classifier Decisions by Layer-Wise Relevance Propagation},
  author={Bach, Sebastian and Binder, Alexander and Montavon, Gr{\'e}goire and Klauschen, Frederick and M{\"u}ller, Klaus-Robert and Samek, Wojciech},
  journal={PLOS ONE},
  year={2015}
}

@inproceedings{zeiler2014visualizing,
  title={Visualizing and Understanding Convolutional Networks},
  author={Zeiler, Matthew D. and Fergus, Rob},
  booktitle={ECCV},
  year={2014}
}

@inproceedings{shrikumar2017learning,
  title={Learning Important Features Through Propagating Activation Differences},
  author={Shrikumar, Avanti and Greenside, Peyton and Kundaje, Anshul},
  booktitle={ICML},
  year={2017}
}

@inproceedings{springenberg2015striving,
  title={Striving for Simplicity: The All Convolutional Net},
  author={Springenberg, Jost Tobias and Dosovitskiy, Alexey and Brox, Thomas and Riedmiller, Martin},
  booktitle={ICLR},
  year={2015}
}

}


\appendix

\section{Appendix}

This appendix provides detailed empirical evidence supporting the alignment analysis presented in the main paper. While the main paper focuses on aggregated findings and decision-level insights, this section exposes the full evaluation pipeline, intermediate results, and statistical validation across all datasets.

Our evaluation framework consists of three key components:
\begin{enumerate}
    \item selection of suitable XAI–ground truth (GT) combinations,
    \item identification of reliable alignment metrics, and
    \item assessment of robustness and stability of the findings.
\end{enumerate}

For each dataset, we report results following a consistent structure to ensure comparability. The appendix includes both soft-metric alignment analysis (continuous attribution-based evaluation) and hard-metric threshold analysis (binarized explanations), together with sensitivity and stability assessments.



The following tables and figures provide complementary perspectives on provided outcomes in the paper.

\subsection{Model Training and Dataset Details}
\label{appx:model_train_dataset}
\begin{table*}[ht]
\centering
\caption{Dataset overview and train/validation/test splits used for predictive model training.}
\label{tab:dataset_overview}
\small
\setlength{\tabcolsep}{5pt}
\begin{adjustbox}{max width=\textwidth}

\begin{tabular}{lllllll}
\toprule
\textbf{Dataset} & \textbf{Task} & \textbf{Classes} & \textbf{Samples} & \textbf{Train/Val/Test} & \textbf{Annotated Test Samples} & \textbf{Input Modality} \\
\midrule
2D Synthetic 
& Multi-label classification 
& 3 (circle, square, triangle) 
& 1400 (balanced) 
& 1120 / 140 / 140 
& 140 (Entire test set)
& RGB images ($224 \times 224$) \\

3D Synthetic 
& Multi-label classification 
& 3 (sphere, cube, pyramid) 
& 1400 (balanced) 
& 980 / 210 / 210 
& 210 (Entire test set)
& 3D volumetric data \\

BRISC2025 
& Multi-class classification 
& 4 
& 6000 T1-weighted MRI slices 
& 5000 / -- / 1000 
& 1000 (Entire test set)
& 2D MRI slices \\

MosMedData 
& Binary classification 
& 2 
& 1110 chest CT scans 
& 777 / -- / 333 
& 50 
& Chest CT volumes \\
\bottomrule
\end{tabular}
\end{adjustbox}
\end{table*}

\begin{table*}[ht]
\centering
\caption{Predictive model architectures and training configurations used for explanation generation.}
\label{tab:model_training}
\small
\setlength{\tabcolsep}{4pt}
\begin{adjustbox}{max width=\textwidth}

\begin{tabular}{llllllll}
\toprule
\textbf{Dataset} & \textbf{Model} & \textbf{Optimizer} & \textbf{LR} & \textbf{Weight Decay} & \textbf{Loss} & \textbf{Epochs} & \textbf{Selection Criterion} \\
\midrule
2D Synthetic 
& Simple CNN with GAP and BatchNorm 
& AdamW 
& $3 \times 10^{-4}$ 
& $1 \times 10^{-4}$ 
& BCEWithLogits 
& 25 
& Best validation per-label accuracy \\

3D Synthetic 
& Basic 3D CNN 
& AdamW 
& $3 \times 10^{-4}$ 
& $1 \times 10^{-4}$ 
& BCEWithLogits 
& 25 
& Per-label and exact-match accuracy \\

BRISC2025 
& DenseNet121 (ImageNet pretrained) 
& Adam 
& $1 \times 10^{-4}$ 
& $1 \times 10^{-4}$ 
& Cross-Entropy 
& 10 
& Best validation macro-F1 \\

MosMedData 
& MONAI 3D DenseNet (pretrained)
& Adam 
& $1 \times 10^{-4}$ 
& $1 \times 10^{-4}$ 
& Cross-Entropy 
& 50 
& Best validation loss \\
\bottomrule
\end{tabular}
\end{adjustbox}
\end{table*}

\subsection{Attribution Heatmaps (Soft Explanations)}
\label{appx:heatmaps}
These figures present representative attribution heatmaps generated by different XAI methods for selected samples. The visualizations are based on normalized soft attribution maps, where higher intensity indicates stronger contribution to the model prediction.

The purpose of these visualizations is to provide qualitative insight into the spatial characteristics of explanations (e.g., region-focused, boundary-focused, or diffuse patterns) and to support the interpretation of quantitative alignment results.

\begin{figure}[h!]
    \centering
    \includegraphics[width=\textwidth]{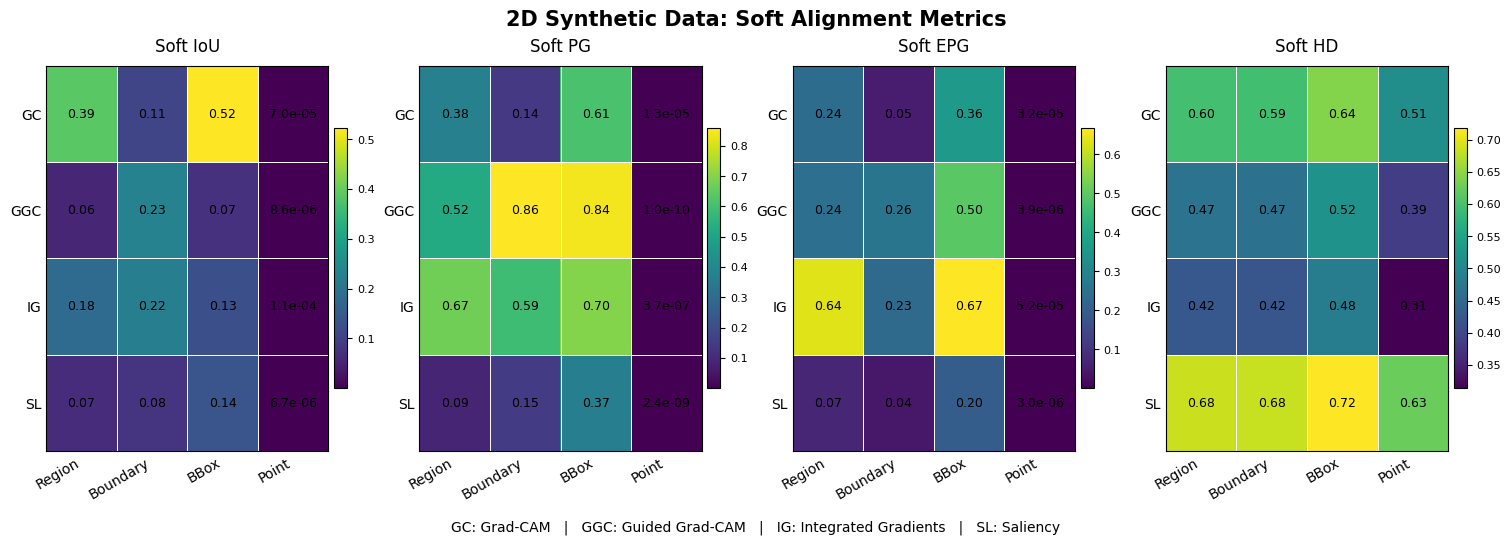}
    \caption{2D Synthetic Data. Heatmaps of XAI Method, Ground Truth Type and Alignment Metric over soft metrics.}
    \label{fig:2DSynt_soft_heatmap}
\end{figure}

\begin{figure}[h!]
    \centering
    \includegraphics[width=\textwidth]{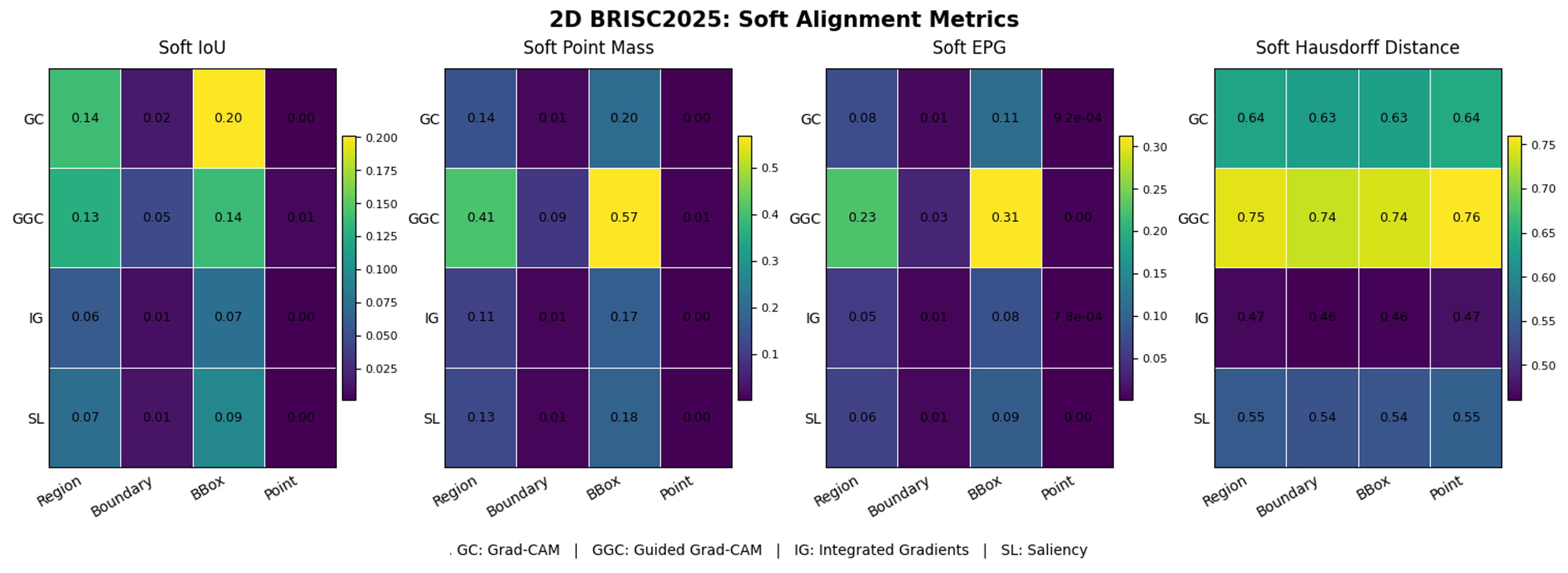}
    \caption{BRISC2025. Heatmaps of XAI Method, Ground Truth Type and Alignment Metric over soft metrics.}
    \label{fig:brisc_soft_heatmap}
\end{figure}

\begin{figure}[h!]
    \centering
    \includegraphics[width=\textwidth]{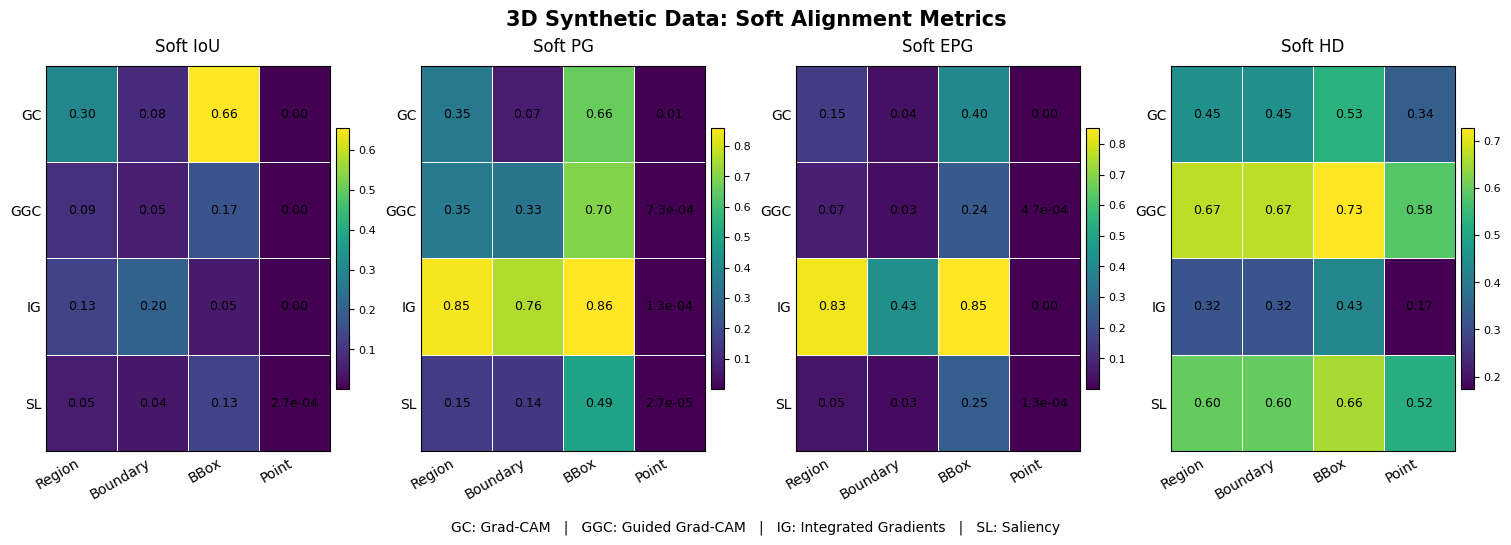}
    \caption{3D Synthetic Data. Heatmaps of XAI Method, Ground Truth Type and Alignment Metric over soft metrics.}
    \label{fig:3DSynt_soft_heatmap}
\end{figure}

\begin{figure}[h!]
    \centering
    \includegraphics[width=\textwidth]{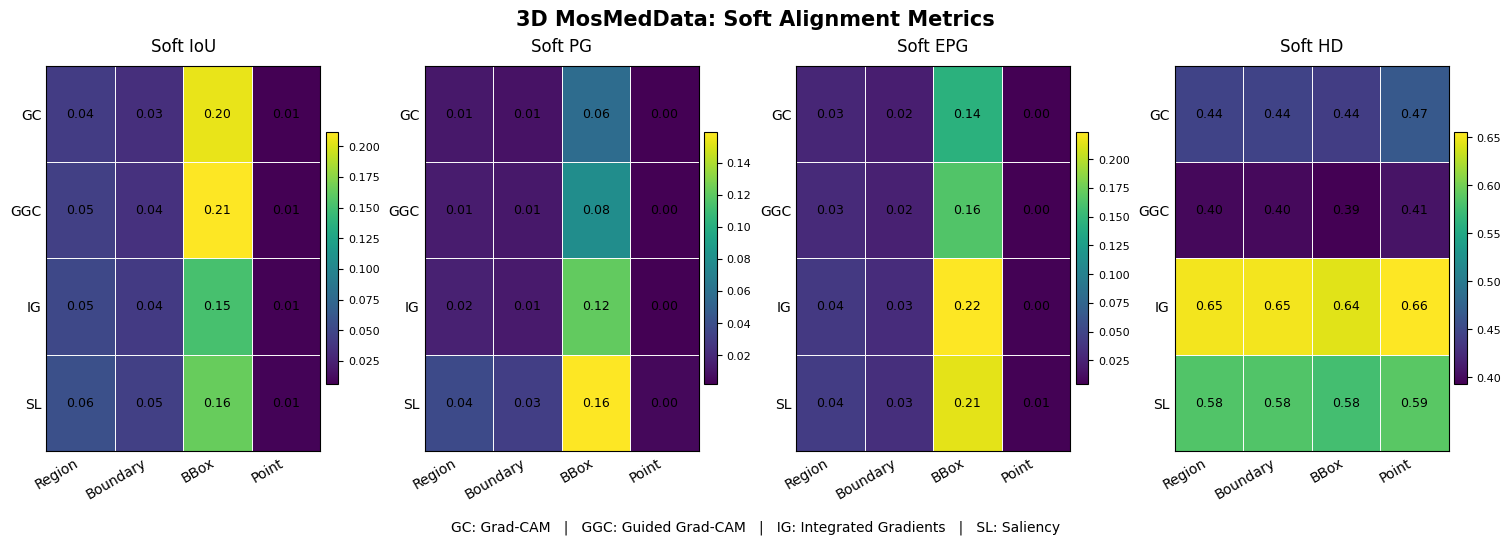}
    \caption{MosMed Data. Heatmaps of XAI Method, Ground Truth Type and Alignment Metric over soft metrics.}
    \label{fig:MosMed_soft_heatmap}
\end{figure}

In particular, these heatmaps (Figure~\ref{fig:2DSynt_soft_heatmap}, \ref{fig:brisc_soft_heatmap}, \ref{fig:3DSynt_soft_heatmap}, \ref{fig:MosMed_soft_heatmap}) illustrate how different XAI methods align with various GT representations and help explain differences observed in metric-based evaluations.

\subsection{Output-Behavior Descriptor Implementation}
\label{appx:descriptor_implementation}

For BMR and LE, the support $S(A)$ retains the locations with the highest
$p=0.10$ fraction of absolute attribution magnitudes; the original
magnitudes are preserved. The inner boundary used by BMR is obtained by
one-step binary erosion with a connectivity-one structuring element.
For LE, we use non-overlapping patches of size $16\times16$ in 2D and
$16\times16\times8$ in 3D, with stride equal to the patch size. Empty
patches and patches with one active location are assigned zero entropy,
and all patches enter the final mean. This fixed support construction is
used only for output characterization and is distinct from thresholding
in the hard alignment evaluation.

\subsection{XAI--GT Candidates (Aggregated Alignment Ranking)}
\label{appx:xai-gt}

These tables summarize the overall alignment quality of each XAI--GT pair under the soft and hard metric formulations. To enable comparison across heterogeneous metrics, lower-is-better HD scores are first direction-aligned and the scores are then standardized per metric using z-score normalization. The reported mean z-score represents the average standardized alignment performance across the four non-redundant metrics retained for preference analysis (IoU, PG, EPG, and HD), while the standard deviation reflects variability across metrics. Rankings are computed within each XAI method to identify the most suitable GT type.

These tables serve as the primary selection mechanism for identifying candidate XAI--GT combinations used in the subsequent metric-level analysis. Tables~\ref{tab:unified_2DSynt_ranking}--\ref{tab:unified_MosMed_ranking} report the complete rankings for all datasets under both formulations.

\begin{table}[t]
\centering
\small
\setlength{\tabcolsep}{6pt}
\caption{2D Synthetic Data. Aggregated alignment rankings across the four non-redundant metrics under soft and hard formulations. GC: Grad-CAM; GGC: Guided Grad-CAM; IG: Integrated Gradients; SL: Saliency. Higher mean z-score indicates better relative alignment across the retained metrics.}
\label{tab:unified_2DSynt_ranking}
\begin{subtable}{.48\textwidth}
\centering
\caption{Soft-metric rankings}
\label{tab:2DSynt_soft_xai_gt_ranking}
\resizebox{\textwidth}{!}{
\begin{tabular}{llccc}
\toprule
Method & GT Type & Mean z-score & Std z-score & Rank \\
\midrule
\rowcolor{groupgray}  & Point & 0.473 & 0.777 & 1 \\
\rowcolor{groupgray}  & BBox & 0.143 & 0.910 & 2 \\
\rowcolor{groupgray}  & Region & 0.133 & 0.848 & 3 \\
\rowcolor{groupgray} \multirow{-4}{*}{GC} & Boundary & -0.693 & 0.183 & 4 \\
\midrule
 & Boundary & 0.931 & 0.236 & 1 \\
 & BBox & 0.343 & 0.746 & 2 \\
 & Region & 0.023 & 0.638 & 3 \\
\multirow{-4}{*}{GGC} & Point & -0.394 & 0.631 & 4 \\
\midrule
\rowcolor{groupgray}  & Region & 0.876 & 0.575 & 1 \\
\rowcolor{groupgray}  & Point & 0.767 & 0.823 & 2 \\
\rowcolor{groupgray}  & Boundary & 0.746 & 0.231 & 3 \\
\rowcolor{groupgray} \multirow{-4}{*}{IG} & BBox & 0.523 & 0.726 & 4 \\
\midrule
 & Point & -0.846 & 0.279 & 1 \\
 & Boundary & -0.984 & 0.167 & 2 \\
 & BBox & -1.009 & 0.426 & 3 \\
\multirow{-4}{*}{SL} & Region & -1.032 & 0.273 & 4 \\
\bottomrule
\end{tabular}}
\end{subtable}
\hfill
\begin{subtable}{.48\textwidth}
\centering
\caption{Hard-metric rankings ($topk=0.1$)}
\label{tab:2DSynt_hard_xai_gt_ranking}
\resizebox{\textwidth}{!}{
\begin{tabular}{llccc}
\toprule
Method & GT Type & Mean z-score & Std z-score & Rank \\
\midrule
\rowcolor{groupgray}  & BBox & 0.552 & 0.541 & 1 \\
\rowcolor{groupgray}  & Region & 0.215 & 0.294 & 2 \\
\rowcolor{groupgray}  & Point & 0.148 & 0.170 & 3 \\
\rowcolor{groupgray} \multirow{-4}{*}{GC} & Boundary & -0.220 & 0.412 & 4 \\
\midrule
 & Boundary & 0.971 & 0.656 & 1 \\
 & BBox & 0.225 & 0.544 & 2 \\
 & Region & -0.097 & 0.536 & 3 \\
\multirow{-4}{*}{GGC} & Point & -0.388 & 0.448 & 4 \\
\midrule
\rowcolor{groupgray}  & Region & 0.780 & 0.540 & 1 \\
\rowcolor{groupgray}  & Point & 0.644 & 0.744 & 2 \\
\rowcolor{groupgray}  & BBox & 0.234 & 0.202 & 3 \\
\rowcolor{groupgray} \multirow{-4}{*}{IG} & Boundary & -0.039 & 0.413 & 4 \\
\midrule
 & Point & -0.404 & 0.466 & 1 \\
 & Boundary & -0.712 & 0.481 & 2 \\
 & Region & -0.898 & 0.613 & 3 \\
\multirow{-4}{*}{SL} & BBox & -1.012 & 0.675 & 4 \\
\bottomrule
\end{tabular}}
\end{subtable}
\end{table}

\begin{table}[]
\centering
\small
\caption{BRISC2025. Aggregated alignment rankings across the four non-redundant metrics under soft and hard formulations. GC: Grad-CAM; GGC: Guided Grad-CAM; IG: Integrated Gradients; SL: Saliency. Higher mean z-score indicates better relative alignment across the retained metrics.}
\label{tab:unified_brisc_ranking}
\begin{subtable}{.48\textwidth}
\centering
\caption{Soft-metric rankings}
\label{tab:brisc_soft_xai_gt_ranking}
\resizebox{\textwidth}{!}{
\begin{tabular}{llccc}
\toprule
Method & GT Type & Mean z-score & Std z-score & Rank \\
\midrule
\rowcolor{groupgray}  & BBox & 0.219 & 0.804 & 1 \\
\rowcolor{groupgray}  & Region & 0.140 & 0.653 & 2 \\
\rowcolor{groupgray}  & Boundary & -0.217 & 0.356 & 3 \\
\rowcolor{groupgray} \multirow{-4}{*}{GC} & Point & -0.434 & 0.547 & 4 \\
\midrule
 & Boundary & 1.427 & 0.133 & 1 \\
 & Point & 1.395 & 0.122 & 2 \\
 & Region & 1.231 & 0.372 & 3 \\
\multirow{-4}{*}{GGC} & BBox & 1.103 & 0.618 & 4 \\
\midrule
\rowcolor{groupgray}  & Point & -0.655 & 0.325 & 1 \\
\rowcolor{groupgray}  & Boundary & -0.715 & 0.260 & 2 \\
\rowcolor{groupgray}  & BBox & -0.807 & 0.246 & 3 \\
\rowcolor{groupgray} \multirow{-4}{*}{IG} & Region & -0.834 & 0.263 & 4 \\
\midrule
 & Point & -0.306 & 0.190 & 1 \\
 & Boundary & -0.496 & 0.036 & 2 \\
 & BBox & -0.515 & 0.058 & 3 \\
\multirow{-4}{*}{SL} & Region & -0.537 & 0.093 & 4 \\
\bottomrule
\end{tabular}}
\end{subtable}
\hfill
\begin{subtable}{.48\textwidth}
\centering
\caption{Hard-metric rankings ($topk=0.1$)}
\label{tab:brisc_hard_xai_gt_ranking}
\resizebox{\textwidth}{!}{
\begin{tabular}{llccc}
\toprule
Method & GT Type & Mean z-score & Std z-score & Rank \\
\midrule
\rowcolor{groupgray}  & Region & 1.212 & 0.205 & 1 \\
\rowcolor{groupgray}  & BBox & 1.206 & 0.210 & 2 \\
\rowcolor{groupgray}  & Boundary & 1.071 & 0.359 & 3 \\
\rowcolor{groupgray} \multirow{-4}{*}{GC} & Point & 0.819 & 0.554 & 4 \\
\midrule
 & Boundary & 0.463 & 0.626 & 1 \\
 & Point & 0.397 & 0.284 & 2 \\
 & BBox & 0.299 & 0.528 & 3 \\
\multirow{-4}{*}{GGC} & Region & 0.287 & 0.537 & 4 \\
\midrule
\rowcolor{groupgray}  & Point & -0.833 & 0.561 & 1 \\
\rowcolor{groupgray}  & Region & -0.860 & 0.241 & 2 \\
\rowcolor{groupgray}  & BBox & -0.879 & 0.244 & 3 \\
\rowcolor{groupgray} \multirow{-4}{*}{IG} & Boundary & -0.892 & 0.228 & 4 \\
\midrule
 & Point & -0.383 & 0.264 & 1 \\
 & BBox & -0.625 & 0.085 & 2 \\
 & Region & -0.639 & 0.107 & 3 \\
\multirow{-4}{*}{SL} & Boundary & -0.642 & 0.058 & 4 \\
\bottomrule
\end{tabular}}
\end{subtable}
\end{table}

\begin{table}[t]
\centering
\small
\setlength{\tabcolsep}{6pt}
\caption{3D Synthetic Data. Aggregated alignment rankings across the four non-redundant metrics under soft and hard formulations. GC: Grad-CAM; GGC: Guided Grad-CAM; IG: Integrated Gradients; SL: Saliency. Higher mean z-score indicates better relative alignment across the retained metrics.}
\label{tab:unified_3DSynt_ranking}
\begin{subtable}{.48\textwidth}
\centering
\caption{Soft-metric rankings}
\label{tab:3DSynt_soft_xai_gt_ranking}
\resizebox{\textwidth}{!}{
\begin{tabular}{llccc}
\toprule
Method & GT Type & Mean z-score & Std z-score & Rank \\
\midrule
\rowcolor{groupgray}  & Point & 0.895 & 0.471 & 1 \\
\rowcolor{groupgray}  & BBox & 0.414 & 0.756 & 2 \\
\rowcolor{groupgray}  & Region & 0.300 & 0.818 & 3 \\
\rowcolor{groupgray} \multirow{-4}{*}{GC} & Boundary & -0.287 & 0.503 & 4 \\
\midrule
 & BBox & -0.479 & 0.523 & 1 \\
 & Boundary & -0.516 & 0.434 & 2 \\
 & Region & -0.581 & 0.333 & 3 \\
\multirow{-4}{*}{GGC} & Point & -0.666 & 0.251 & 4 \\
\midrule
\rowcolor{groupgray}  & Boundary & 1.397 & 0.126 & 1 \\
\rowcolor{groupgray}  & Region & 1.009 & 0.745 & 2 \\
\rowcolor{groupgray}  & BBox & 0.771 & 1.018 & 3 \\
\rowcolor{groupgray} \multirow{-4}{*}{IG} & Point & 0.594 & 0.789 & 4 \\
\midrule
 & Boundary & -0.594 & 0.073 & 1 \\
 & BBox & -0.706 & 0.355 & 2 \\
 & Region & -0.728 & 0.177 & 3 \\
\multirow{-4}{*}{SL} & Point & -0.822 & 0.247 & 4 \\
\bottomrule
\end{tabular}}
\end{subtable}
\hfill
\begin{subtable}{.48\textwidth}
\centering
\caption{Hard-metric rankings ($topk=0.1$)}
\label{tab:3DSynt_hard_xai_gt_ranking}
\resizebox{\textwidth}{!}{
\begin{tabular}{llccc}
\toprule
Method & GT Type & Mean z-score & Std z-score & Rank \\
\midrule
\rowcolor{groupgray}  & Point & 1.313 & 0.291 & 1 \\
\rowcolor{groupgray}  & BBox & 0.921 & 0.291 & 2 \\
\rowcolor{groupgray}  & Region & 0.918 & 0.654 & 3 \\
\rowcolor{groupgray} \multirow{-4}{*}{GC} & Boundary & 0.237 & 0.993 & 4 \\
\midrule
 & Boundary & 0.890 & 0.133 & 1 \\
 & BBox & 0.628 & 0.253 & 2 \\
 & Region & 0.401 & 0.565 & 3 \\
\multirow{-4}{*}{GGC} & Point & -0.220 & 0.598 & 4 \\
\midrule
\rowcolor{groupgray}  & Point & -0.523 & 0.578 & 1 \\
\rowcolor{groupgray}  & Region & -0.556 & 0.670 & 2 \\
\rowcolor{groupgray}  & Boundary & -0.804 & 0.914 & 3 \\
\rowcolor{groupgray} \multirow{-4}{*}{IG} & BBox & -0.922 & 0.503 & 4 \\
\midrule
 & Boundary & -0.323 & 0.175 & 1 \\
 & Point & -0.570 & 0.250 & 2 \\
 & BBox & -0.628 & 0.532 & 3 \\
\multirow{-4}{*}{SL} & Region & -0.764 & 0.524 & 4 \\
\bottomrule
\end{tabular}}
\end{subtable}
\end{table}

\begin{table}[t]
\centering
\small
\setlength{\tabcolsep}{6pt}
\caption{MosMed Data. Aggregated alignment rankings across the four non-redundant metrics under soft and hard formulations. GC: Grad-CAM; GGC: Guided Grad-CAM; IG: Integrated Gradients; SL: Saliency. Higher mean z-score indicates better relative alignment across the retained metrics.}
\label{tab:unified_MosMed_ranking}
\begin{subtable}{.48\textwidth}
\centering
\caption{Soft-metric rankings}
\label{tab:MosMed_soft_xai_gt_ranking}
\resizebox{\textwidth}{!}{
\begin{tabular}{llccc}
\toprule
Method & GT Type & Mean z-score & Std z-score & Rank \\
\midrule
\rowcolor{groupgray}  & BBox & -0.507 & 0.850 & 1 \\
\rowcolor{groupgray}  & Point & -0.554 & 0.062 & 2 \\
\rowcolor{groupgray}  & Boundary & -0.824 & 0.192 & 3 \\
\rowcolor{groupgray} \multirow{-4}{*}{GC} & Region & -0.840 & 0.206 & 4 \\
\midrule
 & BBox & -0.302 & 0.877 & 1 \\
 & Region & -0.691 & 0.238 & 2 \\
 & Boundary & -0.710 & 0.235 & 3 \\
\multirow{-4}{*}{GGC} & Point & -0.913 & 0.330 & 4 \\
\midrule
\rowcolor{groupgray}  & Boundary & 0.469 & 0.611 & 1 \\
\rowcolor{groupgray}  & Region & 0.462 & 0.597 & 2 \\
\rowcolor{groupgray}  & Point & 0.436 & 0.733 & 3 \\
\rowcolor{groupgray} \multirow{-4}{*}{IG} & BBox & 0.350 & 0.987 & 4 \\
\midrule
 & Region & 1.069 & 0.397 & 1 \\
 & Boundary & 1.064 & 0.395 & 2 \\
 & Point & 1.031 & 0.397 & 3 \\
\multirow{-4}{*}{SL} & BBox & 0.460 & 0.790 & 4 \\
\bottomrule
\end{tabular}}
\end{subtable}
\hfill
\begin{subtable}{.48\textwidth}
\centering
\caption{Hard-metric rankings ($topk=0.1$)}
\label{tab:MosMed_hard_xai_gt_ranking}
\resizebox{\textwidth}{!}{
\begin{tabular}{llccc}
\toprule
Method & GT Type & Mean z-score & Std z-score & Rank \\
\midrule
\rowcolor{groupgray}  & Point & 0.486 & 0.562 & 1 \\
\rowcolor{groupgray}  & Boundary & -0.123 & 0.651 & 2 \\
\rowcolor{groupgray}  & Region & -0.374 & 0.488 & 3 \\
\rowcolor{groupgray} \multirow{-4}{*}{GC} & BBox & -0.867 & 0.602 & 4 \\
\midrule
 & Point & -0.023 & 0.028 & 1 \\
 & Boundary & -0.086 & 0.705 & 2 \\
 & Region & -0.262 & 0.498 & 3 \\
\multirow{-4}{*}{GGC} & BBox & -0.291 & 0.275 & 4 \\
\midrule
\rowcolor{groupgray}  & BBox & 0.351 & 0.309 & 1 \\
\rowcolor{groupgray}  & Region & -0.396 & 0.356 & 2 \\
\rowcolor{groupgray}  & Boundary & -0.632 & 0.435 & 3 \\
\rowcolor{groupgray} \multirow{-4}{*}{IG} & Point & -0.682 & 0.787 & 4 \\
\midrule
 & Region & 1.033 & 0.691 & 1 \\
 & Boundary & 0.841 & 0.619 & 2 \\
 & BBox & 0.806 & 0.566 & 3 \\
\multirow{-4}{*}{SL} & Point & 0.219 & 0.254 & 4 \\
\bottomrule
\end{tabular}}
\end{subtable}
\end{table}

\subsubsection{Ranking agreement}
\label{appx:ranking_agreement}
We quantify the differences visible in the flows at the XAI--GT ranking level in Table~\ref{tab:ranking_agreement}. Kendall's $\tau$ summarizes changes in the complete GT ordering, while rank shift and the top-selection measures characterize their magnitude and decision-level consequences.
\begin{table}[H]
\centering
\small
\begin{tabular}{lcccc}
\toprule
\textbf{Dataset} & \textbf{Kendall $\tau$} & \textbf{Rank Shift} & \textbf{Top-1 Change} & \textbf{Top-2 Overlap} \\
\midrule
2D Synthetic   & 0.67  & 0.50 & 0.25 & 0.88 \\
BRISC2025      & 0.42  & 0.75 & 0.25 & 0.75 \\
3D Synthetic   & 0.42  & 0.75 & 0.50 & 0.75 \\
MosMed         & -0.17 & 1.38 & 0.75 & 0.50 \\
\bottomrule
\end{tabular}
\caption{
Ranking agreement between soft and hard metric formulations at the explainable artificial intelligence (XAI)--ground-truth (GT) level after excluding Background Energy Ratio (BER). Kendall's $\tau$ is computed for each XAI method and averaged over each dataset. Rank Shift denotes the mean absolute rank difference, Top-1 Change is the proportion of methods with a different leading GT, and Top-2 Overlap is the mean proportion of shared GT representations in the top-two sets.
}
\label{tab:ranking_agreement}
\end{table}

The 2D synthetic dataset results show the strongest agreement, with $\tau=0.67$ between soft and hard formulations and only one of the four top GT selections changing. BRISC2025 and the 3D synthetic datasets exhibit moderate ordering agreement, although the latter changes the preferred GT for two methods. MosMed dataset shows the strongest formulation dependence: its negative mean $\tau$, largest mean rank shift, and $75\%$ top-1 change rate indicate that thresholding substantially reorders the GT preferences. Nevertheless, the top-2 overlap remains higher than the top-1 agreement in all datasets, showing that formulation changes affect the ordering of plausible GT choices.

\subsection{Metric-Level Rankings for Selected XAI--GT Pairs}
\label{appx:xai-gt-metric}
The triple tables retain the selected pair together with its highest-ranked supporting metrics, and therefore provide the principal dataset-level evidence for both pair and metric preferences. Tables~\ref{tab:unified_2DSynt_metric_support}--\ref{tab:unified_MosMed_metric_support} report the soft and hard results side by side.

\begin{table}[]
\centering
\small
\setlength{\tabcolsep}{5pt}
\caption{2D Synthetic Data. Metric-level support for selected explainable artificial intelligence (XAI)--ground-truth (GT) pairs under soft and hard formulations. GC: Grad-CAM; GGC: Guided Grad-CAM; IG: Integrated Gradients; SL: Saliency; BBox: bounding box; IoU: Intersection over Union; PG: Pointing Game; EPG: Energy-based Pointing Game; HD: Hausdorff Distance; HD$_{dir}$: directed soft HD. All metrics tied within the top two dense ranks are retained.}
\label{tab:unified_2DSynt_metric_support}
\begin{subtable}{.48\textwidth}
\centering
\caption{Soft-metric triples}
\label{tab:2DSynt_soft_metric_support}
\resizebox{\textwidth}{!}{
\begin{tabular}{llccl}
\toprule
Method & GT Type & Rank & Mean z-score & Top supporting metrics \\
\midrule
\rowcolor{groupgray}  & Point & 1 & 0.473 & PG ($z=1.50$); IoU ($z=0.39$) \\
\rowcolor{groupgray} \multirow{-2}{*}{GC} & BBox & 2 & 0.143 & IoU ($z=1.49$); PG ($z=-0.09$) \\
\midrule
 & Boundary & 1 & 0.931 & PG ($z=1.20$); EPG ($z=0.99$) \\
\multirow{-2}{*}{GGC} & BBox & 2 & 0.343 & PG ($z=1.06$); HD$_{dir}$ ($z=0.66$) \\
\midrule
\rowcolor{groupgray}  & Region & 1 & 0.876 & EPG ($z=1.41$); PG ($z=1.03$) \\
\rowcolor{groupgray} \multirow{-2}{*}{IG} & Point & 2 & 0.767 & EPG ($z=1.24$); IoU ($z=1.23$) \\
\midrule
 & Point & 1 & -0.846 & PG ($z=-0.52$); EPG ($z=-0.83$) \\
\multirow{-2}{*}{SL} & Boundary & 2 & -0.984 & PG ($z=-0.81$); EPG ($z=-0.90$) \\
\bottomrule
\end{tabular}}
\end{subtable}
\hfill
\begin{subtable}{.48\textwidth}
\centering
\caption{Hard-metric triples ($topk=0.1$)}
\label{tab:2DSynt_hard_metric_support}
\resizebox{\textwidth}{!}{
\begin{tabular}{llccl}
\toprule
Method & GT Type & Rank & Mean z-score & Top supporting metrics \\
\midrule
\rowcolor{groupgray}  & BBox & 1 & 0.552 & IoU ($z=1.08$); EPG ($z=0.95$) \\
\rowcolor{groupgray} \multirow{-2}{*}{GC} & Region & 2 & 0.215 & EPG, IoU ($z=0.47$) \\
\midrule
 & Boundary & 1 & 0.971 & IoU, EPG ($z=1.37$) \\
\multirow{-2}{*}{GGC} & BBox & 2 & 0.225 & PG ($z=1.04$); HD ($z=0.00$) \\
\midrule
\rowcolor{groupgray}  & Region & 1 & 0.780 & IoU ($z=1.15$); EPG ($z=1.13$) \\
\rowcolor{groupgray} \multirow{-2}{*}{IG} & Point & 2 & 0.644 & IoU, EPG ($z=1.29$); PG, HD ($z=0.00$) \\
\midrule
 & Point & 1 & -0.404 & PG, HD ($z=0.00$); IoU, EPG ($z=-0.81$) \\
\multirow{-2}{*}{SL} & Boundary & 2 & -0.712 & HD ($z=0.00$); PG ($z=-0.84$) \\
\bottomrule
\end{tabular}}
\end{subtable}
\end{table}

\begin{table}[h!]
\centering
\small
\setlength{\tabcolsep}{5pt}
\caption{BRISC2025. Metric-level support for selected explainable artificial intelligence (XAI)--ground-truth (GT) pairs under soft and hard formulations. GC: Grad-CAM; GGC: Guided Grad-CAM; IG: Integrated Gradients; SL: Saliency; BBox: bounding box; IoU: Intersection over Union; PG: Pointing Game; EPG: Energy-based Pointing Game; HD: Hausdorff Distance; HD$_{dir}$: directed soft HD. All metrics tied within the top two dense ranks are retained.}
\label{tab:unified_brisc_metric_support}
\begin{subtable}{.48\textwidth}
\centering
\caption{Soft-metric triples}
\label{tab:brisc_soft_metric_support}
\resizebox{\textwidth}{!}{
\begin{tabular}{llccl}
\toprule
Method & GT Type & Rank & Mean z-score & Top supporting metrics \\
\midrule
\rowcolor{groupgray}  & BBox & 1 & 0.219 & IoU ($z=1.32$); HD$_{dir}$ ($z=0.31$) \\
\rowcolor{groupgray} \multirow{-2}{*}{GC} & Region & 2 & 0.140 & IoU ($z=1.00$); HD$_{dir}$ ($z=0.30$) \\
\midrule
 & Boundary & 1 & 1.427 & PG ($z=1.50$); EPG ($z=1.49$) \\
\multirow{-2}{*}{GGC} & Point & 2 & 1.395 & EPG ($z=1.49$); IoU ($z=1.49$) \\
\midrule
\rowcolor{groupgray}  & Point & 1 & -0.655 & PG ($z=-0.29$); EPG ($z=-0.62$) \\
\rowcolor{groupgray} \multirow{-2}{*}{IG} & Boundary & 2 & -0.715 & PG ($z=-0.51$); EPG ($z=-0.60$) \\
\midrule
 & Point & 1 & -0.306 & PG ($z=-0.03$); IoU ($z=-0.37$) \\
\multirow{-2}{*}{SL} & Boundary & 2 & -0.496 & HD$_{dir}$ ($z=-0.44$); EPG ($z=-0.50$) \\
\bottomrule
\end{tabular}}
\end{subtable}
\hfill
\begin{subtable}{.48\textwidth}
\centering
\caption{Hard-metric triples ($topk=0.1$)}
\label{tab:brisc_hard_metric_support}
\resizebox{\textwidth}{!}{
\begin{tabular}{llccl}
\toprule
Method & GT Type & Rank & Mean z-score & Top supporting metrics \\
\midrule
\rowcolor{groupgray}  & Region & 1 & 1.212 & PG ($z=1.50$); HD ($z=1.21$) \\
\rowcolor{groupgray} \multirow{-2}{*}{GC} & BBox & 2 & 1.206 & PG ($z=1.50$); HD ($z=1.21$) \\
\midrule
 & Boundary & 1 & 0.463 & EPG ($z=0.94$); IoU ($z=0.94$) \\
\multirow{-2}{*}{GGC} & Point & 2 & 0.397 & EPG ($z=0.60$); IoU ($z=0.60$) \\
\midrule
\rowcolor{groupgray}  & Point & 1 & -0.833 & PG ($z=0.00$); HD ($z=-1.00$) \\
\rowcolor{groupgray} \multirow{-2}{*}{IG} & Region & 2 & -0.860 & PG ($z=-0.50$); IoU ($z=-0.97$) \\
\midrule
 & Point & 1 & -0.383 & PG ($z=0.00$); EPG ($z=-0.46$) \\
\multirow{-2}{*}{SL} & BBox & 2 & -0.625 & PG ($z=-0.51$); HD ($z=-0.61$) \\
\bottomrule
\end{tabular}}
\end{subtable}
\end{table}

\begin{table}[]
\centering
\small
\setlength{\tabcolsep}{5pt}
\caption{3D Synthetic Data. Metric-level support for selected explainable artificial intelligence (XAI)--ground-truth (GT) pairs under soft and hard formulations. GC: Grad-CAM; GGC: Guided Grad-CAM; IG: Integrated Gradients; SL: Saliency; BBox: bounding box; IoU: Intersection over Union; PG: Pointing Game; EPG: Energy-based Pointing Game; HD: Hausdorff Distance; HD$_{dir}$: directed soft HD. All metrics tied within the top two dense ranks are retained.}
\label{tab:unified_3DSynt_metric_support}
\begin{subtable}{.48\textwidth}
\centering
\caption{Soft-metric triples}
\label{tab:3DSynt_soft_metric_support}
\resizebox{\textwidth}{!}{
\begin{tabular}{llccl}
\toprule
Method & GT Type & Rank & Mean z-score & Top supporting metrics \\
\midrule
\rowcolor{groupgray}  & Point & 1 & 0.895 & PG ($z=1.49$); IoU ($z=0.92$) \\
\rowcolor{groupgray} \multirow{-2}{*}{GC} & BBox & 2 & 0.414 & IoU ($z=1.47$); HD$_{dir}$ ($z=0.44$) \\
\midrule
 & BBox & 1 & -0.479 & PG ($z=0.14$); IoU ($z=-0.30$) \\
\multirow{-2}{*}{GGC} & Boundary & 2 & -0.516 & PG ($z=0.03$); EPG ($z=-0.50$) \\
\midrule
\rowcolor{groupgray}  & Boundary & 1 & 1.397 & EPG ($z=1.50$); IoU ($z=1.47$) \\
\rowcolor{groupgray} \multirow{-2}{*}{IG} & Region & 2 & 1.009 & EPG ($z=1.49$); PG ($z=1.42$) \\
\midrule
 & Boundary & 1 & -0.594 & EPG ($z=-0.53$); HD$_{dir}$ ($z=-0.55$) \\
\multirow{-2}{*}{SL} & BBox & 2 & -0.706 & IoU ($z=-0.43$); HD$_{dir}$ ($z=-0.53$) \\
\bottomrule
\end{tabular}}
\end{subtable}
\hfill
\begin{subtable}{.48\textwidth}
\centering
\caption{Hard-metric triples ($topk=0.1$)}
\label{tab:3DSynt_hard_metric_support}
\resizebox{\textwidth}{!}{
\begin{tabular}{llccl}
\toprule
Method & GT Type & Rank & Mean z-score & Top supporting metrics \\
\midrule
\rowcolor{groupgray}  & Point & 1 & 1.313 & PG ($z=1.49$); EPG ($z=1.44$) \\
\rowcolor{groupgray} \multirow{-2}{*}{GC} & BBox & 2 & 0.921 & IoU ($z=1.27$); EPG ($z=1.05$) \\
\midrule
 & Boundary & 1 & 0.890 & PG ($z=1.01$); EPG ($z=0.93$) \\
\multirow{-2}{*}{GGC} & BBox & 2 & 0.628 & PG ($z=0.89$); HD ($z=0.77$) \\
\midrule
\rowcolor{groupgray}  & Point & 1 & -0.523 & IoU ($z=-0.10$); EPG ($z=-0.10$) \\
\rowcolor{groupgray} \multirow{-2}{*}{IG} & Region & 2 & -0.556 & PG ($z=0.25$); IoU ($z=-0.49$) \\
\midrule
 & Boundary & 1 & -0.323 & HD ($z=-0.08$); PG ($z=-0.32$) \\
\multirow{-2}{*}{SL} & Point & 2 & -0.570 & HD ($z=-0.21$); PG ($z=-0.57$) \\
\bottomrule
\end{tabular}}
\end{subtable}
\end{table}

\begin{table}[t]
\centering
\small
\setlength{\tabcolsep}{5pt}
\caption{MosMed Data. Metric-level support for selected explainable artificial intelligence (XAI)--ground-truth (GT) pairs under soft and hard formulations. GC: Grad-CAM; GGC: Guided Grad-CAM; IG: Integrated Gradients; SL: Saliency; BBox: bounding box; IoU: Intersection over Union; PG: Pointing Game; EPG: Energy-based Pointing Game; HD: Hausdorff Distance; HD$_{dir}$: directed soft HD. All metrics tied within the top two dense ranks are retained.}
\label{tab:unified_MosMed_metric_support}
\begin{subtable}{.48\textwidth}
\centering
\caption{Soft-metric triples}
\label{tab:MosMed_soft_metric_support}
\resizebox{\textwidth}{!}{
\begin{tabular}{llccl}
\toprule
Method & GT Type & Rank & Mean z-score & Top supporting metrics \\
\midrule
\rowcolor{groupgray}  & BBox & 1 & -0.507 & IoU ($z=0.73$); HD$_{dir}$ ($z=-0.62$) \\
\rowcolor{groupgray} \multirow{-2}{*}{GC} & Point & 2 & -0.554 & PG ($z=-0.47$); HD$_{dir}$ ($z=-0.56$) \\
\midrule
 & BBox & 1 & -0.302 & IoU ($z=0.97$); PG ($z=-0.55$) \\
\multirow{-2}{*}{GGC} & Region & 2 & -0.691 & PG ($z=-0.49$); IoU ($z=-0.56$) \\
\midrule
\rowcolor{groupgray}  & Boundary & 1 & 0.469 & HD$_{dir}$ ($z=1.12$); EPG ($z=0.68$) \\
\rowcolor{groupgray} \multirow{-2}{*}{IG} & Region & 2 & 0.462 & HD$_{dir}$ ($z=1.12$); EPG ($z=0.68$) \\
\midrule
 & Region & 1 & 1.069 & PG ($z=1.48$); IoU ($z=1.23$) \\
\multirow{-2}{*}{SL} & Boundary & 2 & 1.064 & PG ($z=1.48$); IoU ($z=1.20$) \\
\bottomrule
\end{tabular}}
\end{subtable}
\hfill
\begin{subtable}{.48\textwidth}
\centering
\caption{Hard-metric triples ($topk=0.1$)}
\label{tab:MosMed_hard_metric_support}
\resizebox{\textwidth}{!}{
\begin{tabular}{llccl}
\toprule
Method & GT Type & Rank & Mean z-score & Top supporting metrics \\
\midrule
\rowcolor{groupgray}  & Point & 1 & 0.486 & EPG ($z=1.01$); IoU ($z=0.94$) \\
\rowcolor{groupgray} \multirow{-2}{*}{GC} & Boundary & 2 & -0.123 & EPG ($z=0.69$); HD ($z=0.00$) \\
\midrule
 & Point & 1 & -0.023 & PG, HD ($z=0.00$); IoU ($z=-0.03$) \\
\multirow{-2}{*}{GGC} & Boundary & 2 & -0.086 & PG ($z=0.87$); HD ($z=0.00$) \\
\midrule
\rowcolor{groupgray}  & BBox & 1 & 0.351 & EPG ($z=0.65$); IoU ($z=0.56$) \\
\rowcolor{groupgray} \multirow{-2}{*}{IG} & Region & 2 & -0.396 & HD ($z=0.00$); IoU ($z=-0.21$) \\
\midrule
 & Region & 1 & 1.033 & IoU ($z=1.45$); EPG ($z=1.37$) \\
\multirow{-2}{*}{SL} & Boundary & 2 & 0.841 & IoU ($z=1.48$); EPG ($z=1.02$) \\
\bottomrule
\end{tabular}}
\end{subtable}
\end{table}

\subsection{Bootstrap Stability Analysis}
\label{appx:bootstrap}

Tables~\ref{tab:2DSynt_soft_bootstrap_metric_stability}--\ref{tab:MosMed_soft_bootstrap_metric_stability} report the bootstrap confidence intervals and selection frequencies for the soft-metric triples. Pair selection frequency records how often the XAI--GT pair is recovered within the two highest dense ranks, while metric selection frequency records recovery of the supporting metric. Triple confidence is the minimum of these two frequencies.

\begin{table}[]
\centering
\small
\setlength{\tabcolsep}{4pt}
\caption{2D Synthetic Data. Bootstrap stability of the BER-excluded soft-metric triples. Pair and metric selection frequencies denote recovery within the top two dense ranks across 1,000 sample-level resamples; confidence is their minimum.}
\label{tab:2DSynt_soft_bootstrap_metric_stability}
\resizebox{\textwidth}{!}{%
\begin{tabular}{llccc lccc}
\toprule
Method & GT Type & Mean z-score & 95\% CI & GT sel. & Metric & Metric sel. & Metric z-score & Conf. \\
\midrule
\rowcolor{groupgray}  &  &  &  &  & PG & 1.000 & 1.499 & High \\
\rowcolor{groupgray}  & \multirow{-2}{*}{Point} & \multirow{-2}{*}{0.474} & \multirow{-2}{*}{[0.429, 0.522]} & \multirow{-2}{*}{1.000} & IoU & 1.000 & 0.395 & High \\
\rowcolor{groupgray}  &  &  &  &  & IoU & 1.000 & 1.487 & High \\
\rowcolor{groupgray} \multirow{-4}{*}{GC} & \multirow{-2}{*}{BBox} & \multirow{-2}{*}{0.143} & \multirow{-2}{*}{[0.112, 0.169]} & \multirow{-2}{*}{0.970} & PG & 1.000 & -0.089 & High \\
\midrule
 &  &  &  &  & PG & 1.000 & 1.203 & High \\
 & \multirow{-2}{*}{Boundary} & \multirow{-2}{*}{0.930} & \multirow{-2}{*}{[0.891, 0.965]} & \multirow{-2}{*}{1.000} & EPG & 1.000 & 0.984 & High \\
 &  &  &  &  & PG & 1.000 & 1.059 & High \\
\multirow{-4}{*}{GGC} & \multirow{-2}{*}{BBox} & \multirow{-2}{*}{0.342} & \multirow{-2}{*}{[0.314, 0.368]} & \multirow{-2}{*}{1.000} & HD$_{dir}$ & 1.000 & 0.659 & High \\
\midrule
\rowcolor{groupgray}  &  &  &  &  & EPG & 1.000 & 1.415 & High \\
\rowcolor{groupgray}  & \multirow{-2}{*}{Region} & \multirow{-2}{*}{0.876} & \multirow{-2}{*}{[0.855, 0.899]} & \multirow{-2}{*}{1.000} & PG & 0.977 & 1.033 & High \\
\rowcolor{groupgray}  &  &  &  &  & EPG & 1.000 & 1.233 & High \\
\rowcolor{groupgray} \multirow{-4}{*}{IG} & \multirow{-2}{*}{Point} & \multirow{-2}{*}{0.766} & \multirow{-2}{*}{[0.734, 0.795]} & \multirow{-2}{*}{0.835} & IoU & 1.000 & 1.232 & High \\
\midrule
 &  &  &  &  & PG & 1.000 & -0.519 & High \\
 & \multirow{-2}{*}{Point} & \multirow{-2}{*}{-0.846} & \multirow{-2}{*}{[-0.860, -0.830]} & \multirow{-2}{*}{1.000} & EPG & 1.000 & -0.831 & High \\
 &  &  &  &  & PG & 1.000 & -0.805 & High \\
\multirow{-4}{*}{SL} & \multirow{-2}{*}{Boundary} & \multirow{-2}{*}{-0.984} & \multirow{-2}{*}{[-0.995, -0.972]} & \multirow{-2}{*}{1.000} & EPG & 1.000 & -0.897 & High \\
\bottomrule
\end{tabular}}
\end{table}

\begin{table}[]
\centering
\small
\setlength{\tabcolsep}{4pt}
\caption{BRISC2025. Bootstrap stability of the BER-excluded soft-metric triples. Pair and metric selection frequencies denote recovery within the top two dense ranks across 1,000 sample-level resamples; confidence is their minimum.}
\label{tab:brisc_soft_bootstrap_metric_stability}
\resizebox{\textwidth}{!}{%
\begin{tabular}{llccc lccc}
\toprule
Method & GT Type & Mean z-score & 95\% CI & GT sel. & Metric & Metric sel. & Metric z-score & Conf. \\
\midrule
\rowcolor{groupgray}  &  &  &  &  & IoU & 1.000 & 1.320 & High \\
\rowcolor{groupgray}  & \multirow{-2}{*}{BBox} & \multirow{-2}{*}{0.219} & \multirow{-2}{*}{[0.192, 0.245]} & \multirow{-2}{*}{1.000} & HD$_{dir}$ & 1.000 & 0.305 & High \\
\rowcolor{groupgray}  &  &  &  &  & IoU & 1.000 & 0.992 & High \\
\rowcolor{groupgray} \multirow{-4}{*}{GC} & \multirow{-2}{*}{Region} & \multirow{-2}{*}{0.139} & \multirow{-2}{*}{[0.095, 0.177]} & \multirow{-2}{*}{1.000} & HD$_{dir}$ & 1.000 & 0.301 & High \\
\midrule
 &  &  &  &  & PG & 1.000 & 1.499 & High \\
 & \multirow{-2}{*}{Boundary} & \multirow{-2}{*}{1.427} & \multirow{-2}{*}{[1.424, 1.430]} & \multirow{-2}{*}{1.000} & EPG & 1.000 & 1.495 & High \\
 &  &  &  &  & EPG & 1.000 & 1.492 & High \\
\multirow{-4}{*}{GGC} & \multirow{-2}{*}{Point} & \multirow{-2}{*}{1.384} & \multirow{-2}{*}{[1.289, 1.423]} & \multirow{-2}{*}{0.997} & IoU & 1.000 & 1.492 & High \\
\midrule
\rowcolor{groupgray}  &  &  &  &  & PG & 0.994 & -0.266 & High \\
\rowcolor{groupgray}  & \multirow{-2}{*}{Point} & \multirow{-2}{*}{-0.648} & \multirow{-2}{*}{[-0.738, -0.523]} & \multirow{-2}{*}{1.000} & EPG & 1.000 & -0.618 & High \\
\rowcolor{groupgray}  &  &  &  &  & PG & 1.000 & -0.511 & High \\
\rowcolor{groupgray} \multirow{-4}{*}{IG} & \multirow{-2}{*}{Boundary} & \multirow{-2}{*}{-0.715} & \multirow{-2}{*}{[-0.729, -0.701]} & \multirow{-2}{*}{1.000} & EPG & 1.000 & -0.600 & High \\
\midrule
 &  &  &  &  & PG & 0.991 & -0.005 & High \\
 & \multirow{-2}{*}{Point} & \multirow{-2}{*}{-0.301} & \multirow{-2}{*}{[-0.401, -0.168]} & \multirow{-2}{*}{1.000} & IoU & 1.000 & -0.372 & High \\
 &  &  &  &  & HD$_{dir}$ & 1.000 & -0.444 & High \\
\multirow{-4}{*}{SL} & \multirow{-2}{*}{Boundary} & \multirow{-2}{*}{-0.496} & \multirow{-2}{*}{[-0.508, -0.484]} & \multirow{-2}{*}{0.960} & EPG & 0.992 & -0.500 & High \\
\bottomrule
\end{tabular}}
\end{table}

\begin{table}[]
\centering
\small
\setlength{\tabcolsep}{4pt}
\caption{3D Synthetic Data. Bootstrap stability of the BER-excluded soft-metric triples. Pair and metric selection frequencies denote recovery within the top two dense ranks across 1,000 sample-level resamples; confidence is their minimum.}
\label{tab:3DSynt_soft_bootstrap_metric_stability}
\resizebox{\textwidth}{!}{%
\begin{tabular}{llccc lccc}
\toprule
Method & GT Type & Mean z-score & 95\% CI & GT sel. & Metric & Metric sel. & Metric z-score & Conf. \\
\midrule
\rowcolor{groupgray}  &  &  &  &  & PG & 1.000 & 1.485 & High \\
\rowcolor{groupgray}  & \multirow{-2}{*}{Point} & \multirow{-2}{*}{0.894} & \multirow{-2}{*}{[0.836, 0.948]} & \multirow{-2}{*}{1.000} & IoU & 1.000 & 0.918 & High \\
\rowcolor{groupgray}  &  &  &  &  & IoU & 1.000 & 1.473 & High \\
\rowcolor{groupgray} \multirow{-4}{*}{GC} & \multirow{-2}{*}{BBox} & \multirow{-2}{*}{0.413} & \multirow{-2}{*}{[0.380, 0.447]} & \multirow{-2}{*}{1.000} & HD$_{dir}$ & 1.000 & 0.443 & High \\
\midrule
 &  &  &  &  & PG & 1.000 & 0.143 & High \\
 & \multirow{-2}{*}{BBox} & \multirow{-2}{*}{-0.480} & \multirow{-2}{*}{[-0.501, -0.458]} & \multirow{-2}{*}{1.000} & IoU & 1.000 & -0.296 & High \\
 &  &  &  &  & PG & 1.000 & 0.027 & High \\
\multirow{-4}{*}{GGC} & \multirow{-2}{*}{Boundary} & \multirow{-2}{*}{-0.516} & \multirow{-2}{*}{[-0.530, -0.500]} & \multirow{-2}{*}{1.000} & EPG & 1.000 & -0.501 & High \\
\midrule
\rowcolor{groupgray}  &  &  &  &  & EPG & 1.000 & 1.500 & High \\
\rowcolor{groupgray}  & \multirow{-2}{*}{Boundary} & \multirow{-2}{*}{1.397} & \multirow{-2}{*}{[1.391, 1.403]} & \multirow{-2}{*}{1.000} & IoU & 1.000 & 1.470 & High \\
\rowcolor{groupgray}  &  &  &  &  & EPG & 1.000 & 1.490 & High \\
\rowcolor{groupgray} \multirow{-4}{*}{IG} & \multirow{-2}{*}{Region} & \multirow{-2}{*}{1.009} & \multirow{-2}{*}{[0.993, 1.024]} & \multirow{-2}{*}{1.000} & PG & 1.000 & 1.421 & High \\
\midrule
 &  &  &  &  & EPG & 1.000 & -0.528 & High \\
 & \multirow{-2}{*}{Boundary} & \multirow{-2}{*}{-0.594} & \multirow{-2}{*}{[-0.604, -0.585]} & \multirow{-2}{*}{1.000} & HD$_{dir}$ & 1.000 & -0.554 & High \\
 &  &  &  &  & IoU & 1.000 & -0.434 & High \\
\multirow{-4}{*}{SL} & \multirow{-2}{*}{BBox} & \multirow{-2}{*}{-0.705} & \multirow{-2}{*}{[-0.722, -0.687]} & \multirow{-2}{*}{1.000} & HD$_{dir}$ & 1.000 & -0.532 & High \\
\bottomrule
\end{tabular}}
\end{table}

\begin{table}[]
\centering
\small
\setlength{\tabcolsep}{4pt}
\caption{MosMed Data. Bootstrap stability of the BER-excluded soft-metric triples. Pair and metric selection frequencies denote recovery within the top two dense ranks across 1,000 sample-level resamples; confidence is their minimum.}
\label{tab:MosMed_soft_bootstrap_metric_stability}
\resizebox{\textwidth}{!}{%
\begin{tabular}{llccc lccc}
\toprule
Method & GT Type & Mean z-score & 95\% CI & GT sel. & Metric & Metric sel. & Metric z-score & Conf. \\
\midrule
\rowcolor{groupgray}  &  &  &  &  & IoU & 1.000 & 0.637 & High \\
\rowcolor{groupgray}  & \multirow{-2}{*}{BBox} & \multirow{-2}{*}{-0.509} & \multirow{-2}{*}{[-0.881, -0.125]} & \multirow{-2}{*}{0.988} & HD$_{dir}$ & 0.976 & -0.621 & High \\
\rowcolor{groupgray}  &  &  &  &  & PG & 0.766 & -0.377 & High \\
\rowcolor{groupgray} \multirow{-4}{*}{GC} & \multirow{-2}{*}{Point} & \multirow{-2}{*}{-0.510} & \multirow{-2}{*}{[-0.859, -0.015]} & \multirow{-2}{*}{0.970} & HD$_{dir}$ & 0.542 & -0.557 & Medium \\
\midrule
 &  &  &  &  & IoU & 1.000 & 0.932 & High \\
 & \multirow{-2}{*}{BBox} & \multirow{-2}{*}{-0.312} & \multirow{-2}{*}{[-0.665, -0.046]} & \multirow{-2}{*}{1.000} & PG & 0.601 & -0.536 & Medium \\
 &  &  &  &  & PG & 0.923 & -0.462 & Medium \\
\multirow{-4}{*}{GGC} & \multirow{-2}{*}{Region} & \multirow{-2}{*}{-0.692} & \multirow{-2}{*}{[-1.000, -0.443]} & \multirow{-2}{*}{0.742} & IoU & 0.800 & -0.587 & Medium \\
\midrule
\rowcolor{groupgray}  &  &  &  &  & HD$_{dir}$ & 1.000 & 1.112 & High \\
\rowcolor{groupgray}  & \multirow{-2}{*}{Boundary} & \multirow{-2}{*}{0.446} & \multirow{-2}{*}{[0.181, 0.593]} & \multirow{-2}{*}{0.816} & EPG & 1.000 & 0.670 & High \\
\rowcolor{groupgray}  &  &  &  &  & HD$_{dir}$ & 1.000 & 1.112 & Medium \\
\rowcolor{groupgray} \multirow{-4}{*}{IG} & \multirow{-2}{*}{Region} & \multirow{-2}{*}{0.437} & \multirow{-2}{*}{[0.151, 0.590]} & \multirow{-2}{*}{0.648} & EPG & 1.000 & 0.673 & Medium \\
\midrule
 &  &  &  &  & PG & 0.994 & 1.454 & High \\
 & \multirow{-2}{*}{Region} & \multirow{-2}{*}{1.054} & \multirow{-2}{*}{[0.963, 1.143]} & \multirow{-2}{*}{0.891} & IoU & 0.993 & 1.209 & High \\
 &  &  &  &  & PG & 0.993 & 1.458 & High \\
\multirow{-4}{*}{SL} & \multirow{-2}{*}{Boundary} & \multirow{-2}{*}{1.048} & \multirow{-2}{*}{[0.962, 1.134]} & \multirow{-2}{*}{0.891} & IoU & 0.993 & 1.177 & High \\
\bottomrule
\end{tabular}}
\end{table}

\subsection{Threshold Sensitivity Plots (Hard Metrics)}
\label{appx:sensitivity}
These figures visualize the behavior of hard alignment metrics across different threshold levels. Each plot shows the mean metric value across samples, with shaded regions representing standard deviation. The dashed vertical line indicates the estimated stability onset, and the gray shaded region represents the common stable interval when it exists across all methods.

In cases where no common stable threshold exists, this is explicitly indicated in the plot.

These visualizations complement the tabular summaries by providing an intuitive view of metric sensitivity and stability.

\begin{figure}[!htbp]
    \centering
    \includegraphics[width=\textwidth]{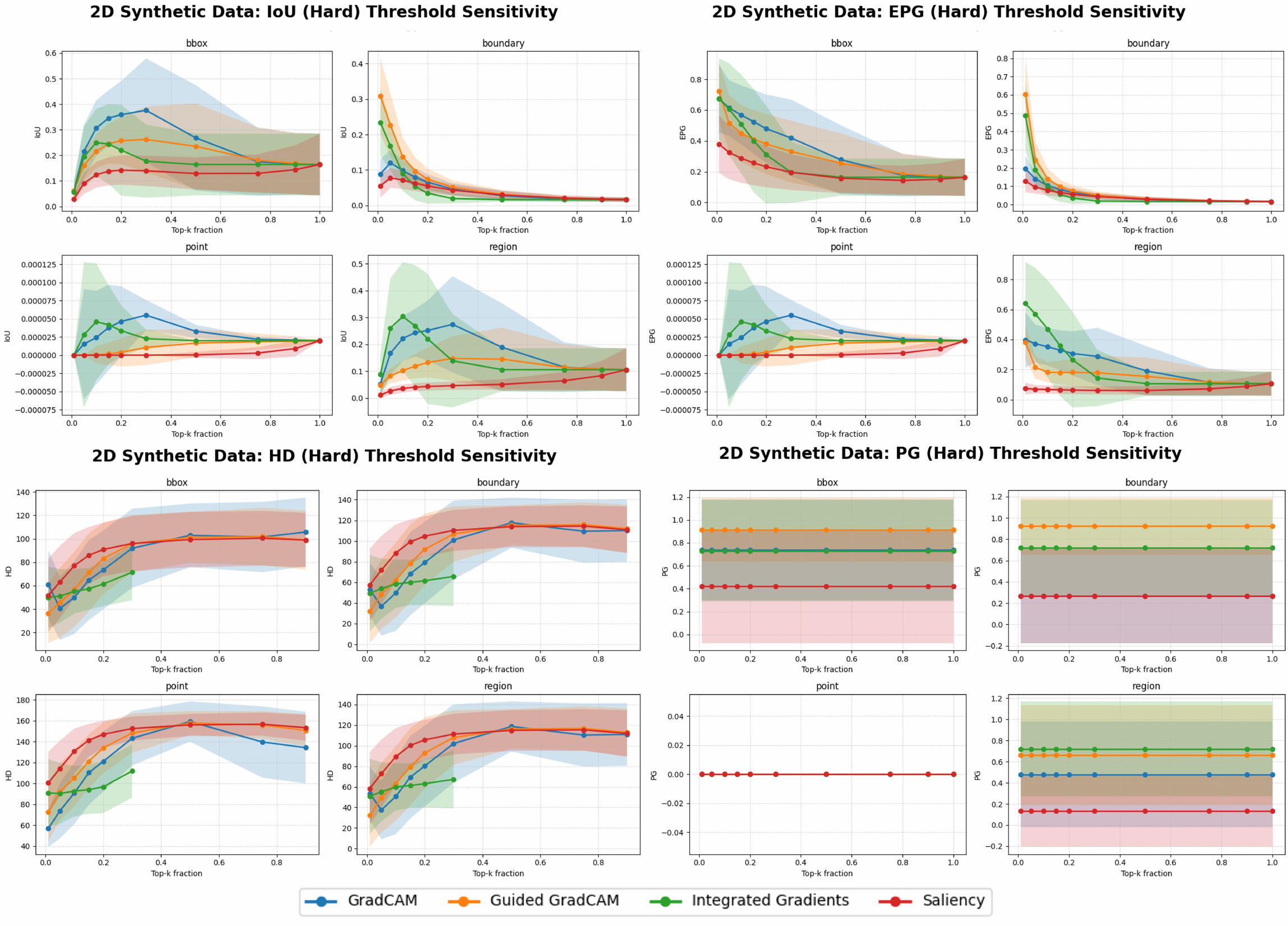}
    \caption{2D Synthetic Data. Hard-metric threshold sensitivity across GT types.}
    \label{fig:2DSynt_hard_sensitivity}
\end{figure}

\begin{figure}[!htbp]
    \centering
    \includegraphics[width=\textwidth]{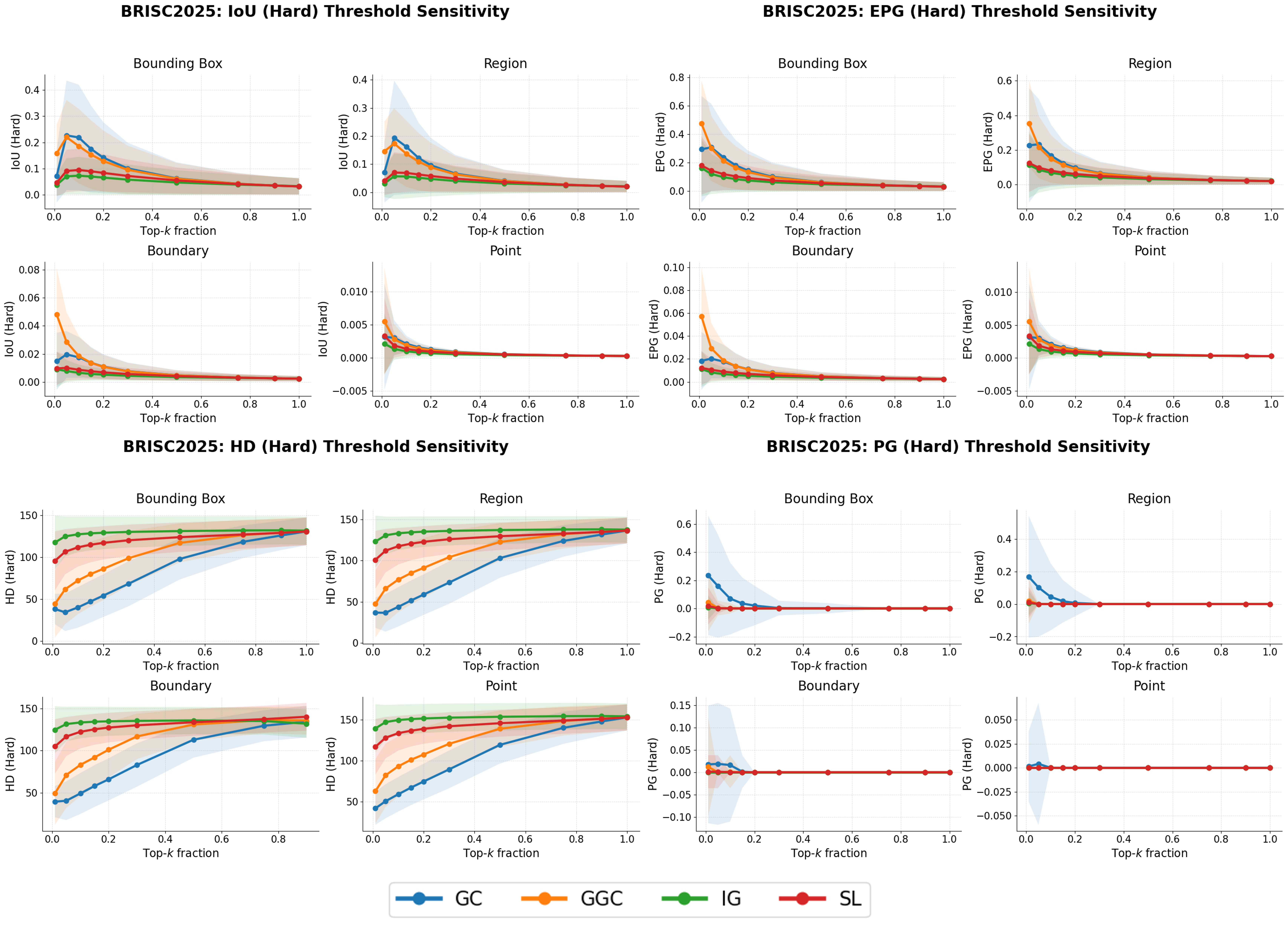}
    \caption{BRISC2025. Hard-metric threshold sensitivity across GT types.}
    \label{fig:brisc_hard_sensitivity}
\end{figure}

\begin{figure}[!htbp]
    \centering
    \includegraphics[width=\textwidth]{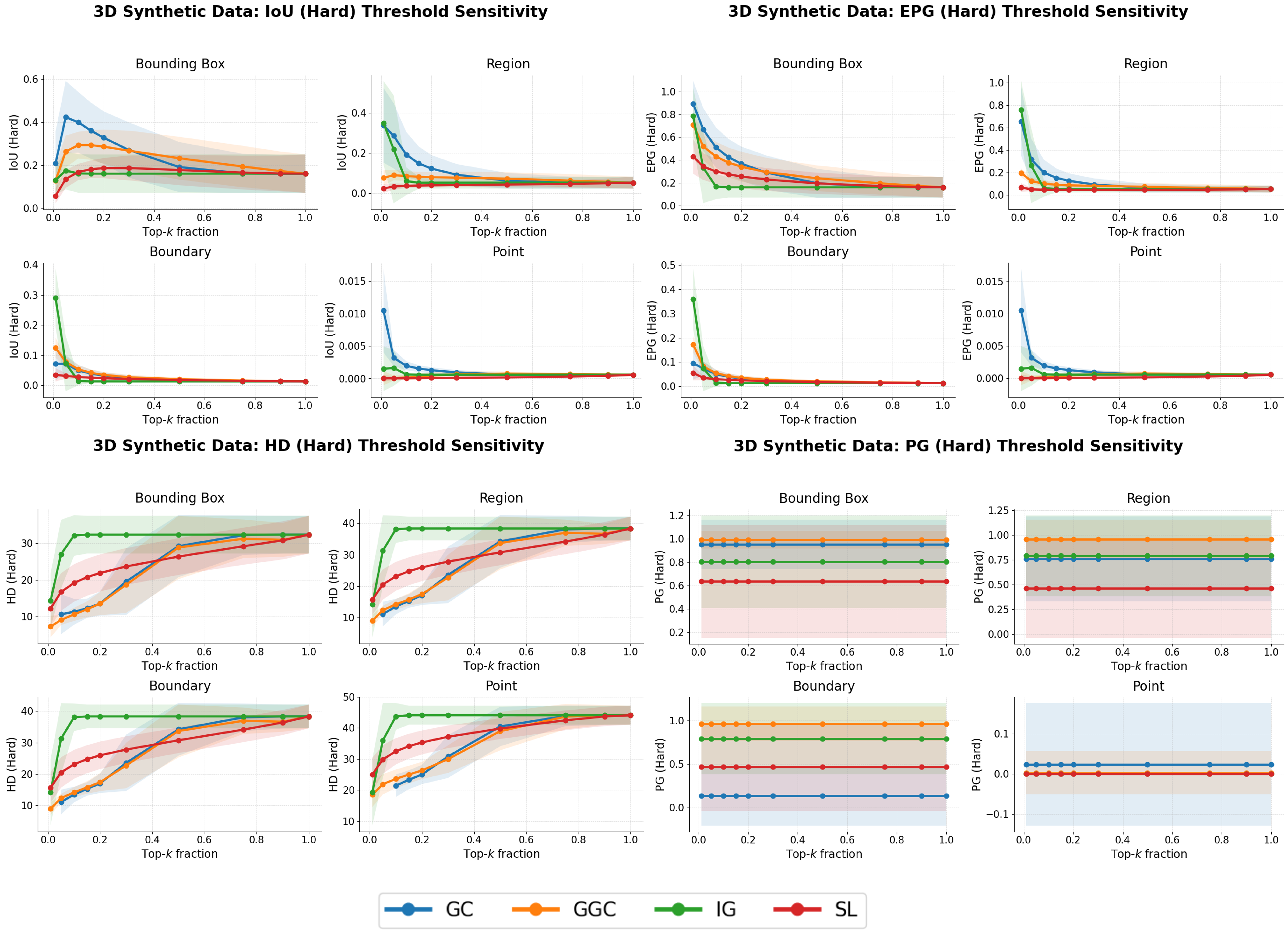}
    \caption{3D Synthetic Data. Hard-metric threshold sensitivity across GT types.}
    \label{fig:3DSynt_hard_sensitivity}
\end{figure}

\begin{figure}[!htbp]
    \centering
    \includegraphics[width=\textwidth]{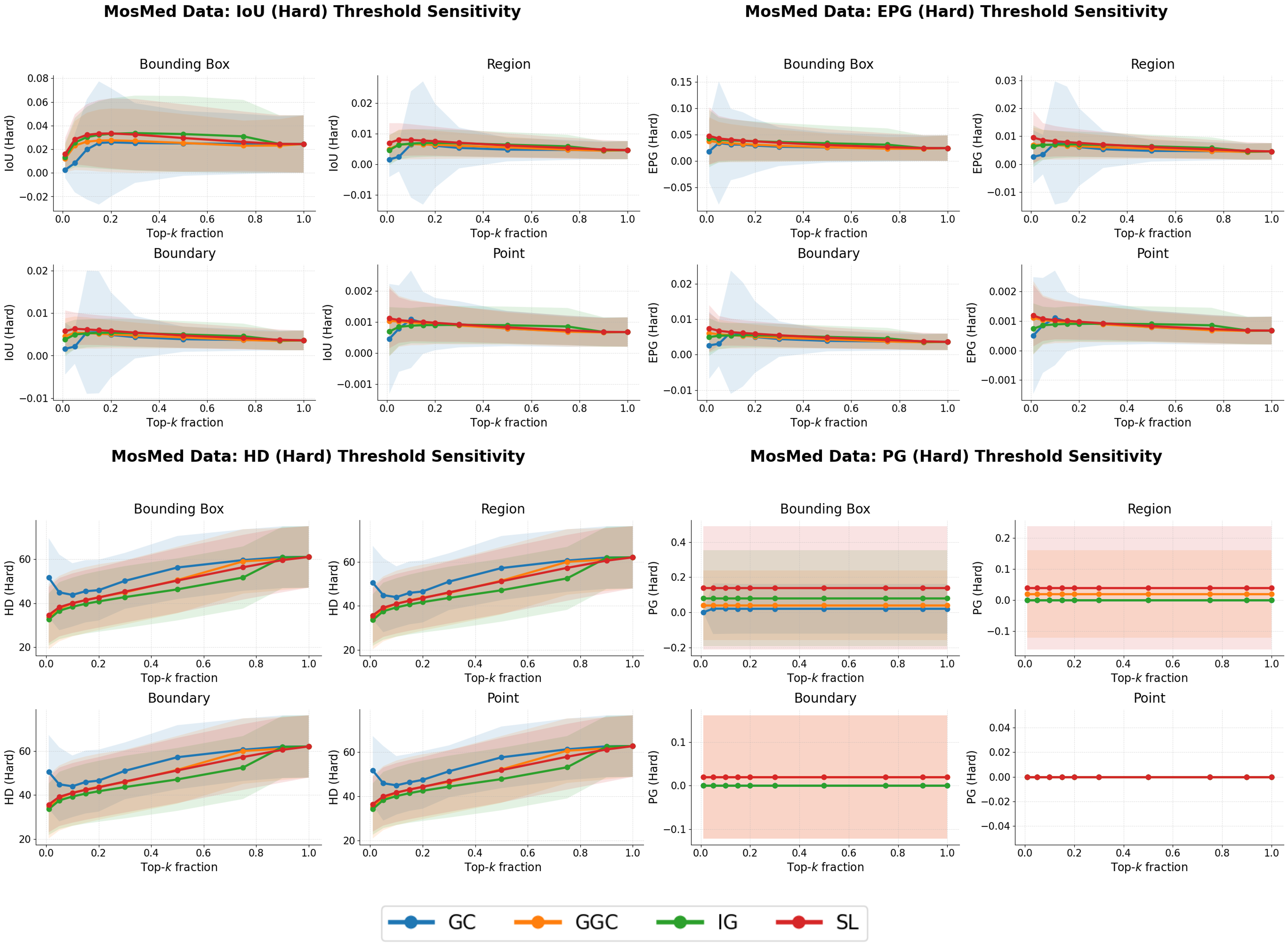}
    \caption{MosMed Data. Hard-metric threshold sensitivity across GT types.}
    \label{fig:MosMed_hard_sensitivity}
\end{figure}

Figures~\ref{fig:2DSynt_hard_sensitivity}, \ref{fig:3DSynt_hard_sensitivity}, \ref{fig:brisc_hard_sensitivity}, \ref{fig:MosMed_hard_sensitivity} show metric values as a function of the top-$k$ threshold for four XAI methods: GC (Grad-CAM), GGC (Guided Grad-CAM), IG (Integrated Gradients), and SL (Saliency). Shaded regions indicate variability across samples. Dashed vertical lines mark the common stability onset, and gray regions denote the corresponding stable threshold interval, computed using a slope-based stability criterion. When not shown, a common stability region does not exist because at least one method does not exhibit stable behavior for the given metric and GT type.

\subsection{Hard-Metric Threshold Sensitivity \& Stability Tables}
\label{appx:hard-sensitivity-stability}

These tables characterize the behavior of hard (threshold-based) alignment metrics under varying top-$k$ thresholds. For each method–GT–metric combination, we compute mean alignment curves across thresholds and derive sensitivity and stability measures from these curves.

We report the following quantities:

\begin{itemize}

\item \textbf{Stability onset:} the smallest threshold at which the metric enters its \emph{global plateau}. The global plateau is defined as the largest contiguous region near the end of the threshold curve where metric values remain within a tolerance band around the final value. Specifically, a point is considered stable if its deviation from the final value is less than a fraction $\tau$ of the total dynamic range of the curve (here $\tau = 0.05$). The stability onset corresponds to the beginning of this plateau region.

\item \textbf{Stability interval:} the continuous range of thresholds from the stability onset to the maximum threshold (i.e., $1.0$), representing the global plateau where the metric has effectively converged.

\item \textbf{Threshold sensitivity:} the overall variability of the metric across thresholds, quantified using the interquartile range (IQR) of the aligned metric values. Let $Q_1$ and $Q_2$ denote the first and second tertiles (i.e., $33^{\text{rd}}$ and $66^{\text{th}}$ percentiles) computed over all combinations. Sensitivity is assigned as:
\[
\text{Low} \;\; \text{if} \;\; \text{IQR} \leq Q_1,\quad
\text{Medium} \;\; \text{if} \;\; Q_1 < \text{IQR} \leq Q_2,\quad
\text{High} \;\; \text{if} \;\; \text{IQR} > Q_2,
\]
where lower IQR indicates lower sensitivity (i.e., more robust behavior with respect to threshold selection).

\item \textbf{Stability:} the proportion of thresholds that fall within the global plateau (denoted as the \emph{stable fraction}). Using the same tertile thresholds $Q_1$ and $Q_2$ computed over all combinations, stability is assigned as:
\[
\text{Low} \;\; \text{if} \;\; f \leq Q_1,\quad
\text{Medium} \;\; \text{if} \;\; Q_1 < f \leq Q_2,\quad
\text{High} \;\; \text{if} \;\; f > Q_2,
\]
where $f$ denotes the stable fraction, and higher values indicate more consistent convergence across thresholds.

\end{itemize}

For interpretability, both threshold sensitivity and stability are discretized into \textit{Low}, \textit{Medium}, and \textit{High} categories using tertile-based thresholds across all combinations. Sensitivity labels reflect how strongly metric values vary across thresholds, while stability labels capture how consistently the metric converges to a plateau.

These measures provide a principled way to assess the reliability of hard metrics under threshold selection, highlighting whether conclusions drawn from a specific threshold (e.g., top-$10\%$) are stable or sensitive to this choice.

\small
\begin{longtable}{lllcccc}
\caption{2D Synthetic Dataset. Hard-metric threshold sensitivity and stability summary. GC: Grad-CAM; GGC: Guided Grad-CAM; IG: Integrated Gradients; SL: Saliency. HD denotes Hausdorff Distance. ``--'' indicates that no stable threshold region was detected.}
\label{tab:2DSynt_hard_threshold_stability}\\
\toprule
Method & GT Type & Metric & Stability onset & Stability interval & Sensitivity Robustness& Stability \\
\midrule
\endfirsthead
\toprule
Method & GT Type & Metric & Stability onset & Stability interval & Sensitivity Robustness & Stability \\
\midrule
\endhead
\midrule
\multicolumn{7}{r}{Continued on next page} \\
\endfoot
\bottomrule
\endlastfoot

\rowcolor{groupgray}  &  & BER & 0.90 & [0.90, 1.00] & \low & \low \\*
\rowcolor{groupgray}  &  & EPG & 0.90 & [0.90, 1.00] & \low & \low \\*
\rowcolor{groupgray}  &  & HD & 0.75 & [0.75, 0.90] & \low & \medium \\*
\rowcolor{groupgray}  &  & IoU & 0.90 & [0.90, 1.00] & \low & \low \\*
\rowcolor{groupgray}  & \multirow{-5}{*}{BBox} & PG & 0.01 & [0.01, 1.00] & \high & \high \\
\rowcolor{groupgray}  &  & BER & 0.90 & [0.90, 1.00] & \low & \low \\*
\rowcolor{groupgray}  &  & EPG & 0.90 & [0.90, 1.00] & \low & \low \\*
\rowcolor{groupgray}  &  & HD & 0.50 & [0.50, 0.90] & \low & \medium \\*
\rowcolor{groupgray}  &  & IoU & 0.90 & [0.90, 1.00] & \medium & \low \\*
\rowcolor{groupgray}  & \multirow{-5}{*}{Region} & PG & 0.01 & [0.01, 1.00] & \high & \high \\
\rowcolor{groupgray}  &  & BER & 0.75 & [0.75, 1.00] & \medium & \medium \\*
\rowcolor{groupgray}  &  & EPG & 0.75 & [0.75, 1.00] & \medium & \medium \\*
\rowcolor{groupgray}  &  & HD & 0.50 & [0.50, 0.90] & \low & \medium \\*
\rowcolor{groupgray}  &  & IoU & 0.90 & [0.90, 1.00] & \medium & \low \\*
\rowcolor{groupgray}  & \multirow{-5}{*}{Boundary} & PG & 0.01 & [0.01, 1.00] & \high & \high \\
\rowcolor{groupgray}  &  & BER & 0.90 & [0.90, 1.00] & \high & \low \\*
\rowcolor{groupgray}  &  & EPG & 0.90 & [0.90, 1.00] & \high & \low \\*
\rowcolor{groupgray}  &  & HD & -- & -- & \low & \low \\*
\rowcolor{groupgray}  &  & IoU & 0.90 & [0.90, 1.00] & \high & \low \\*
\rowcolor{groupgray} \multirow{-20}{*}{GC} & \multirow{-5}{*}{Point} & PG & 0.01 & [0.01, 1.00] & \high & \high \\
\midrule

 &  & BER & 0.90 & [0.90, 1.00] & \low & \low \\*
 &  & EPG & 0.90 & [0.90, 1.00] & \low & \low \\*
 &  & HD & 0.50 & [0.50, 0.90] & \low & \medium \\*
 &  & IoU & 0.90 & [0.90, 1.00] & \medium & \low \\*
 & \multirow{-5}{*}{BBox} & PG & 0.01 & [0.01, 1.00] & \high & \high \\
 &  & BER & 0.90 & [0.90, 1.00] & \medium & \low \\*
 &  & EPG & 0.90 & [0.90, 1.00] & \medium & \low \\*
 &  & HD & 0.50 & [0.50, 0.90] & \low & \medium \\*
 &  & IoU & 0.90 & [0.90, 1.00] & \medium & \low \\*
 & \multirow{-5}{*}{Region} & PG & 0.01 & [0.01, 1.00] & \high & \high \\
 &  & BER & 0.50 & [0.50, 1.00] & \medium & \high \\*
 &  & EPG & 0.50 & [0.50, 1.00] & \medium & \high \\*
 &  & HD & 0.50 & [0.50, 0.90] & \low & \medium \\*
 &  & IoU & 0.75 & [0.75, 1.00] & \medium & \medium \\*
 & \multirow{-5}{*}{Boundary} & PG & 0.01 & [0.01, 1.00] & \high & \high \\
 &  & BER & 0.90 & [0.90, 1.00] & \high & \low \\*
 &  & EPG & 0.90 & [0.90, 1.00] & \high & \low \\*
 &  & HD & 0.50 & [0.50, 0.90] & \low & \medium \\*
 &  & IoU & 0.90 & [0.90, 1.00] & \high & \low \\*
\multirow{-20}{*}{GGC} & \multirow{-5}{*}{Point} & PG & 0.01 & [0.01, 1.00] & \high & \high \\
\midrule

\rowcolor{groupgray}  &  & BER & 0.50 & [0.50, 1.00] & \low & \high \\*
\rowcolor{groupgray}  &  & EPG & 0.50 & [0.50, 1.00] & \low & \high \\*
\rowcolor{groupgray}  &  & HD & -- & -- & \low & \low \\*
\rowcolor{groupgray}  &  & IoU & 0.75 & [0.75, 1.00] & \medium & \medium \\*
\rowcolor{groupgray}  & \multirow{-5}{*}{BBox} & PG & 0.01 & [0.01, 1.00] & \high & \high \\
\rowcolor{groupgray}  &  & BER & 0.50 & [0.50, 1.00] & \low & \high \\*
\rowcolor{groupgray}  &  & EPG & 0.50 & [0.50, 1.00] & \low & \high \\*
\rowcolor{groupgray}  &  & HD & -- & -- & \low & \low \\*
\rowcolor{groupgray}  &  & IoU & 0.75 & [0.75, 1.00] & \medium & \medium \\*
\rowcolor{groupgray}  & \multirow{-5}{*}{Region} & PG & 0.01 & [0.01, 1.00] & \high & \high \\
\rowcolor{groupgray}  &  & BER & 0.30 & [0.30, 1.00] & \medium & \high \\*
\rowcolor{groupgray}  &  & EPG & 0.30 & [0.30, 1.00] & \medium & \high \\*
\rowcolor{groupgray}  &  & HD & -- & -- & \low & \low \\*
\rowcolor{groupgray}  &  & IoU & 0.30 & [0.30, 1.00] & \medium & \high \\*
\rowcolor{groupgray}  & \multirow{-5}{*}{Boundary} & PG & 0.01 & [0.01, 1.00] & \high & \high \\
\rowcolor{groupgray}  &  & BER & 0.75 & [0.75, 1.00] & \high & \medium \\*
\rowcolor{groupgray}  &  & EPG & 0.75 & [0.75, 1.00] & \high & \medium \\*
\rowcolor{groupgray}  &  & HD & -- & -- & \low & \low \\*
\rowcolor{groupgray}  &  & IoU & 0.75 & [0.75, 1.00] & \high & \medium \\*
\rowcolor{groupgray} \multirow{-20}{*}{IG} & \multirow{-5}{*}{Point} & PG & 0.01 & [0.01, 1.00] & \high & \high \\
\midrule

 &  & BER & 0.50 & [0.50, 1.00] & \medium & \high \\*
 &  & EPG & 0.50 & [0.50, 1.00] & \medium & \high \\*
 &  & HD & 0.50 & [0.50, 0.90] & \low & \medium \\*
 &  & IoU & -- & -- & \medium & \low \\*
 & \multirow{-5}{*}{BBox} & PG & 0.01 & [0.01, 1.00] & \high & \high \\
 &  & BER & -- & -- & \medium & \low \\*
 &  & EPG & -- & -- & \medium & \low \\*
 &  & HD & 0.50 & [0.50, 0.90] & \low & \medium \\*
 &  & IoU & -- & -- & \medium & \low \\*
 & \multirow{-5}{*}{Region} & PG & 0.01 & [0.01, 1.00] & \high & \high \\
 &  & BER & 0.90 & [0.90, 1.00] & \medium & \low \\*
 &  & EPG & 0.90 & [0.90, 1.00] & \medium & \low \\*
 &  & HD & 0.50 & [0.50, 0.90] & \low & \medium \\*
 &  & IoU & 0.90 & [0.90, 1.00] & \medium & \low \\*
 & \multirow{-5}{*}{Boundary} & PG & 0.01 & [0.01, 1.00] & \high & \high \\
 &  & BER & -- & -- & \high & \low \\*
 &  & EPG & -- & -- & \high & \low \\*
 &  & HD & 0.50 & [0.50, 0.90] & \low & \medium \\*
 &  & IoU & -- & -- & \high & \low \\*
\multirow{-20}{*}{SL} & \multirow{-5}{*}{Point} & PG & 0.01 & [0.01, 1.00] & \high & \high \\
\end{longtable}

\newpage
\small
\begin{longtable}{lllcccc}
\caption{BRISC2025. Hard-metric threshold sensitivity and stability summary. GC: Grad-CAM; GGC: Guided Grad-CAM; IG: Integrated Gradients; SL: Saliency. HD denotes Hausdorff Distance. ``--'' indicates that no stable threshold region was detected.}\label{tab:brisc_hard_threshold_stability}\\
\toprule
Method & GT Type & Metric & Stability onset & Stability interval & Sensitivity Robustness & Stability \\
\midrule
\endfirsthead
\toprule
Method & GT Type & Metric & Stability onset & Stability interval & Sensitivity Robustness & Stability \\
\midrule
\endhead
\midrule
\multicolumn{7}{r}{Continued on next page} \\
\endfoot
\bottomrule
\endlastfoot

\rowcolor{groupgray}  &  & BER & 0.90 & [0.90, 1.00] & \low & \low \\*
\rowcolor{groupgray}  &  & EPG & 0.90 & [0.90, 1.00] & \low & \low \\*
\rowcolor{groupgray}  &  & HD & -- & -- & \low & \low \\*
\rowcolor{groupgray}  &  & IoU & 0.90 & [0.90, 1.00] & \low & \low \\*
\rowcolor{groupgray}  & \multirow{-5}{*}{BBox} & PG & 0.30 & [0.30, 1.00] & \medium & \high \\
\rowcolor{groupgray}  &  & BER & 0.75 & [0.75, 1.00] & \low & \medium \\*
\rowcolor{groupgray}  &  & EPG & 0.75 & [0.75, 1.00] & \low & \medium \\*
\rowcolor{groupgray}  &  & HD & 0.90 & [0.90, 1.00] & \low & \low \\*
\rowcolor{groupgray}  &  & IoU & 0.90 & [0.90, 1.00] & \medium & \low \\*
\rowcolor{groupgray}  & \multirow{-5}{*}{Region} & PG & 0.30 & [0.30, 1.00] & \medium & \high \\
\rowcolor{groupgray}  &  & BER & 0.90 & [0.90, 1.00] & \medium & \low \\*
\rowcolor{groupgray}  &  & EPG & 0.90 & [0.90, 1.00] & \medium & \low \\*
\rowcolor{groupgray}  &  & HD & -- & -- & \low & \low \\*
\rowcolor{groupgray}  &  & IoU & 0.90 & [0.90, 1.00] & \medium & \low \\*
\rowcolor{groupgray}  & \multirow{-5}{*}{Boundary} & PG & 0.30 & [0.30, 1.00] & \medium & \high \\
\rowcolor{groupgray}  &  & BER & 0.75 & [0.75, 1.00] & \high & \medium \\*
\rowcolor{groupgray}  &  & EPG & 0.75 & [0.75, 1.00] & \high & \medium \\*
\rowcolor{groupgray}  &  & HD & 0.90 & [0.90, 1.00] & \low & \low \\*
\rowcolor{groupgray}  &  & IoU & 0.75 & [0.75, 1.00] & \high & \medium \\*
\rowcolor{groupgray} \multirow{-20}{*}{GC} & \multirow{-5}{*}{Point} & PG & 0.20 & [0.20, 1.00] & \high & \high \\
\midrule

 &  & BER & 0.75 & [0.75, 1.00] & \low & \medium \\*
 &  & EPG & 0.75 & [0.75, 1.00] & \low & \medium \\*
 &  & HD & 0.90 & [0.90, 1.00] & \low & \low \\*
 &  & IoU & 0.90 & [0.90, 1.00] & \low & \low \\*
 & \multirow{-5}{*}{BBox} & PG & 0.15 & [0.15, 1.00] & \high & \high \\
 &  & BER & 0.75 & [0.75, 1.00] & \low & \medium \\*
 &  & EPG & 0.75 & [0.75, 1.00] & \low & \medium \\*
 &  & HD & 0.90 & [0.90, 1.00] & \low & \low \\*
 &  & IoU & 0.90 & [0.90, 1.00] & \low & \low \\*
 & \multirow{-5}{*}{Region} & PG & 0.15 & [0.15, 1.00] & \high & \high \\
 &  & BER & 0.75 & [0.75, 1.00] & \medium & \medium \\*
 &  & EPG & 0.75 & [0.75, 1.00] & \medium & \medium \\*
 &  & HD & 0.75 & [0.75, 0.90] & \low & \medium \\*
 &  & IoU & 0.75 & [0.75, 1.00] & \medium & \medium \\*
 & \multirow{-5}{*}{Boundary} & PG & 0.15 & [0.15, 1.00] & \high & \high \\
 &  & BER & 0.75 & [0.75, 1.00] & \high & \medium \\*
 &  & EPG & 0.75 & [0.75, 1.00] & \high & \medium \\*
 &  & HD & 0.90 & [0.90, 1.00] & \low & \low \\*
 &  & IoU & 0.75 & [0.75, 1.00] & \high & \medium \\*
\multirow{-20}{*}{GGC} & \multirow{-5}{*}{Point} & PG & 0.01 & [0.01, 1.00] & \high & \high \\
\midrule

\rowcolor{groupgray}  &  & BER & 0.90 & [0.90, 1.00] & \medium & \low \\*
\rowcolor{groupgray}  &  & EPG & 0.90 & [0.90, 1.00] & \medium & \low \\*
\rowcolor{groupgray}  &  & HD & 0.75 & [0.75, 1.00] & \low & \medium \\*
\rowcolor{groupgray}  &  & IoU & -- & -- & \medium & \low \\*
\rowcolor{groupgray}  & \multirow{-5}{*}{BBox} & PG & 0.15 & [0.15, 1.00] & \high & \high \\
\rowcolor{groupgray}  &  & BER & 0.90 & [0.90, 1.00] & \medium & \low \\*
\rowcolor{groupgray}  &  & EPG & 0.90 & [0.90, 1.00] & \medium & \low \\*
\rowcolor{groupgray}  &  & HD & 0.75 & [0.75, 1.00] & \low & \medium \\*
\rowcolor{groupgray}  &  & IoU & 0.90 & [0.90, 1.00] & \medium & \low \\*
\rowcolor{groupgray}  & \multirow{-5}{*}{Region} & PG & 0.15 & [0.15, 1.00] & \high & \high \\
\rowcolor{groupgray}  &  & BER & 0.90 & [0.90, 1.00] & \high & \low \\*
\rowcolor{groupgray}  &  & EPG & 0.90 & [0.90, 1.00] & \medium & \low \\*
\rowcolor{groupgray}  &  & HD & -- & -- & \low & \low \\*
\rowcolor{groupgray}  &  & IoU & 0.90 & [0.90, 1.00] & \high & \low \\*
\rowcolor{groupgray}  & \multirow{-5}{*}{Boundary} & PG & 0.01 & [0.01, 1.00] & \high & \high \\
\rowcolor{groupgray}  &  & BER & 0.75 & [0.75, 1.00] & \high & \medium \\*
\rowcolor{groupgray}  &  & EPG & 0.75 & [0.75, 1.00] & \high & \medium \\*
\rowcolor{groupgray}  &  & HD & 0.75 & [0.75, 1.00] & \low & \medium \\*
\rowcolor{groupgray}  &  & IoU & 0.75 & [0.75, 1.00] & \high & \medium \\*
\rowcolor{groupgray} \multirow{-20}{*}{IG} & \multirow{-5}{*}{Point} & PG & 0.01 & [0.01, 1.00] & \high & \high \\
\midrule

 &  & BER & 0.90 & [0.90, 1.00] & \medium & \low \\*
 &  & EPG & 0.90 & [0.90, 1.00] & \medium & \low \\*
 &  & HD & 0.90 & [0.90, 1.00] & \low & \low \\*
 &  & IoU & -- & -- & \medium & \low \\*
 & \multirow{-5}{*}{BBox} & PG & 0.15 & [0.15, 1.00] & \high & \high \\
 &  & BER & 0.90 & [0.90, 1.00] & \medium & \low \\*
 &  & EPG & 0.90 & [0.90, 1.00] & \medium & \low \\*
 &  & HD & 0.90 & [0.90, 1.00] & \low & \low \\*
 &  & IoU & -- & -- & \medium & \low \\*
 & \multirow{-5}{*}{Region} & PG & 0.15 & [0.15, 1.00] & \high & \high \\
 &  & BER & 0.90 & [0.90, 1.00] & \medium & \low \\*
 &  & EPG & 0.90 & [0.90, 1.00] & \medium & \low \\*
 &  & HD & -- & -- & \low & \low \\*
 &  & IoU & 0.90 & [0.90, 1.00] & \medium & \low \\*
 & \multirow{-5}{*}{Boundary} & PG & 0.20 & [0.20, 1.00] & \high & \high \\
 &  & BER & 0.75 & [0.75, 1.00] & \high & \medium \\*
 &  & EPG & 0.75 & [0.75, 1.00] & \high & \medium \\*
 &  & HD & 0.90 & [0.90, 1.00] & \low & \low \\*
 &  & IoU & 0.75 & [0.75, 1.00] & \high & \medium \\*
\multirow{-20}{*}{SL} & \multirow{-5}{*}{Point} & PG & 0.01 & [0.01, 1.00] & \high & \high \\
\end{longtable}

\begin{longtable}{lllcccc}
\caption{3D Synthetic Data. Hard-metric threshold sensitivity and stability summary. GC: Grad-CAM; GGC: Guided Grad-CAM; IG: Integrated Gradients; SL: Saliency. HD denotes Hausdorff Distance. ``--'' indicates that no stable threshold region was detected.}
\label{tab:3DSynt_hard_threshold_stability}\\
\toprule
Method & GT Type & Metric & Stability onset & Stability interval & Sensitivity Robustness & Stability \\
\midrule
\endfirsthead
\toprule
Method & GT Type & Metric & Stability onset & Stability interval & Sensitivity Robustness & Stability \\
\midrule
\endhead
\midrule
\multicolumn{7}{r}{Continued on next page} \\
\endfoot
\bottomrule
\endlastfoot

\rowcolor{groupgray}  &  & BER & 0.50 & [0.50, 1.00] & \low & \low \\*
\rowcolor{groupgray}  &  & EPG & 0.50 & [0.50, 1.00] & \low & \low \\*
\rowcolor{groupgray}  &  & HD & 0.90 & [0.90, 1.00] & \low & \low \\*
\rowcolor{groupgray}  &  & IoU & 0.75 & [0.75, 1.00] & \low & \low \\*
\rowcolor{groupgray}  & \multirow{-5}{*}{BBox} & PG & 0.01 & [0.01, 1.00] & \high & \high \\
\rowcolor{groupgray}  &  & BER & 0.30 & [0.30, 1.00] & \low & \medium \\*
\rowcolor{groupgray}  &  & EPG & 0.30 & [0.30, 1.00] & \low & \medium \\*
\rowcolor{groupgray}  &  & HD & 0.90 & [0.90, 1.00] & \low & \low \\*
\rowcolor{groupgray}  &  & IoU & 0.50 & [0.50, 1.00] & \low & \low \\*
\rowcolor{groupgray}  & \multirow{-5}{*}{Region} & PG & 0.01 & [0.01, 1.00] & \high & \high \\
\rowcolor{groupgray}  &  & BER & 0.50 & [0.50, 1.00] & \medium & \low \\*
\rowcolor{groupgray}  &  & EPG & 0.50 & [0.50, 1.00] & \medium & \low \\*
\rowcolor{groupgray}  &  & HD & 0.90 & [0.90, 1.00] & \low & \low \\*
\rowcolor{groupgray}  &  & IoU & 0.75 & [0.75, 1.00] & \medium & \low \\*
\rowcolor{groupgray}  & \multirow{-5}{*}{Boundary} & PG & 0.01 & [0.01, 1.00] & \high & \high \\
\rowcolor{groupgray}  &  & BER & 0.20 & [0.20, 1.00] & \medium & \medium \\*
\rowcolor{groupgray}  &  & EPG & 0.20 & [0.20, 1.00] & \medium & \medium \\*
\rowcolor{groupgray}  &  & HD & 0.90 & [0.90, 1.00] & \low & \low \\*
\rowcolor{groupgray}  &  & IoU & 0.20 & [0.20, 1.00] & \high & \medium \\*
\rowcolor{groupgray} \multirow{-20}{*}{GC} & \multirow{-5}{*}{Point} & PG & 0.01 & [0.01, 1.00] & \high & \high \\
\midrule

 &  & BER & 0.50 & [0.50, 1.00] & \low & \low \\*
 &  & EPG & 0.50 & [0.50, 1.00] & \low & \low \\*
 &  & HD & -- & -- & \low & \low \\*
 &  & IoU & 0.90 & [0.90, 1.00] & \low & \low \\*
 & \multirow{-5}{*}{BBox} & PG & 0.01 & [0.01, 1.00] & \high & \high \\
 &  & BER & 0.30 & [0.30, 1.00] & \medium & \medium \\*
 &  & EPG & 0.30 & [0.30, 1.00] & \medium & \medium \\*
 &  & HD & -- & -- & \low & \low \\*
 &  & IoU & -- & -- & \medium & \low \\*
 & \multirow{-5}{*}{Region} & PG & 0.01 & [0.01, 1.00] & \high & \high \\
 &  & BER & 0.30 & [0.30, 1.00] & \medium & \medium \\*
 &  & EPG & 0.30 & [0.30, 1.00] & \medium & \medium \\*
 &  & HD & -- & -- & \low & \low \\*
 &  & IoU & 0.50 & [0.50, 1.00] & \medium & \low \\*
 & \multirow{-5}{*}{Boundary} & PG & 0.01 & [0.01, 1.00] & \high & \high \\
 &  & BER & 0.50 & [0.50, 0.75] & \high & \low \\*
 &  & EPG & 0.50 & [0.50, 0.75] & \high & \low \\*
 &  & HD & -- & -- & \low & \low \\*
 &  & IoU & 0.50 & [0.50, 0.75] & \high & \low \\*
\multirow{-20}{*}{GGC} & \multirow{-5}{*}{Point} & PG & 0.01 & [0.01, 1.00] & \high & \high \\
\midrule

\rowcolor{groupgray}  &  & BER & 0.20 & [0.20, 1.00] & \medium & \medium \\*
\rowcolor{groupgray}  &  & EPG & 0.20 & [0.20, 1.00] & \medium & \medium \\*
\rowcolor{groupgray}  &  & HD & 0.20 & [0.20, 1.00] & \low & \medium \\*
\rowcolor{groupgray}  &  & IoU & 0.15 & [0.15, 1.00] & \high & \high \\*
\rowcolor{groupgray}  & \multirow{-5}{*}{BBox} & PG & 0.01 & [0.01, 1.00] & \high & \high \\
\rowcolor{groupgray}  &  & BER & 0.20 & [0.20, 1.00] & \medium & \medium \\*
\rowcolor{groupgray}  &  & EPG & 0.20 & [0.20, 1.00] & \medium & \medium \\*
\rowcolor{groupgray}  &  & HD & 0.20 & [0.20, 1.00] & \low & \medium \\*
\rowcolor{groupgray}  &  & IoU & 0.20 & [0.20, 1.00] & \medium & \medium \\*
\rowcolor{groupgray}  & \multirow{-5}{*}{Region} & PG & 0.01 & [0.01, 1.00] & \high & \high \\
\rowcolor{groupgray}  &  & BER & 0.20 & [0.20, 1.00] & \medium & \medium \\*
\rowcolor{groupgray}  &  & EPG & 0.20 & [0.20, 1.00] & \medium & \medium \\*
\rowcolor{groupgray}  &  & HD & 0.20 & [0.20, 1.00] & \low & \medium \\*
\rowcolor{groupgray}  &  & IoU & 0.20 & [0.20, 1.00] & \medium & \medium \\*
\rowcolor{groupgray}  & \multirow{-5}{*}{Boundary} & PG & 0.01 & [0.01, 1.00] & \high & \high \\
\rowcolor{groupgray}  &  & BER & 0.30 & [0.30, 1.00] & \high & \medium \\*
\rowcolor{groupgray}  &  & EPG & 0.30 & [0.30, 1.00] & \high & \medium \\*
\rowcolor{groupgray}  &  & HD & 0.20 & [0.20, 1.00] & \low & \medium \\*
\rowcolor{groupgray}  &  & IoU & 0.30 & [0.30, 1.00] & \high & \medium \\*
\rowcolor{groupgray} \multirow{-20}{*}{IG} & \multirow{-5}{*}{Point} & PG & 0.01 & [0.01, 1.00] & \high & \high \\
\midrule

 &  & BER & 0.75 & [0.75, 1.00] & \low & \low \\*
 &  & EPG & 0.75 & [0.75, 1.00] & \low & \low \\*
 &  & HD & -- & -- & \low & \low \\*
 &  & IoU & 0.20 & [0.20, 1.00] & \medium & \medium \\*
 & \multirow{-5}{*}{BBox} & PG & 0.01 & [0.01, 1.00] & \high & \high \\
 &  & BER & 0.20 & [0.20, 1.00] & \medium & \medium \\*
 &  & EPG & 0.20 & [0.20, 1.00] & \medium & \medium \\*
 &  & HD & -- & -- & \low & \low \\*
 &  & IoU & -- & -- & \medium & \low \\*
 & \multirow{-5}{*}{Region} & PG & 0.01 & [0.01, 1.00] & \high & \high \\
 &  & BER & 0.50 & [0.50, 1.00] & \medium & \low \\*
 &  & EPG & 0.50 & [0.50, 1.00] & \medium & \low \\*
 &  & HD & -- & -- & \low & \low \\*
 &  & IoU & -- & -- & \medium & \low \\*
 & \multirow{-5}{*}{Boundary} & PG & 0.01 & [0.01, 1.00] & \high & \high \\
 &  & BER & -- & -- & \high & \low \\*
 &  & EPG & -- & -- & \high & \low \\*
 &  & HD & -- & -- & \low & \low \\*
 &  & IoU & -- & -- & \high & \low \\*
\multirow{-20}{*}{SL} & \multirow{-5}{*}{Point} & PG & 0.01 & [0.01, 1.00] & \high & \high \\
\end{longtable}

\begin{longtable}{lllcccc}
\caption{MosMed Data. Hard-metric threshold sensitivity and stability summary. GC: Grad-CAM; GGC: Guided Grad-CAM; IG: Integrated Gradients; SL: Saliency. HD denotes Hausdorff Distance. ``--'' indicates that no stable threshold region was detected.}
\label{tab:MosMed_hard_threshold_stability}\\
\toprule
Method & GT Type & Metric & Stability onset & Stability interval & Sensitivity Robustness & Stability \\
\midrule
\endfirsthead
\toprule
Method & GT Type & Metric & Stability onset & Stability interval & Sensitivity Robustness & Stability \\
\midrule
\endhead
\midrule
\multicolumn{7}{r}{Continued on next page} \\
\endfoot
\bottomrule
\endlastfoot

\rowcolor{groupgray}  &  & BER & 0.90 & [0.90, 1.00] & \low & \medium \\*
\rowcolor{groupgray}  &  & EPG & 0.90 & [0.90, 1.00] & \low & \medium \\*
\rowcolor{groupgray}  &  & HD & 0.90 & [0.90, 1.00] & \low & \medium \\*
\rowcolor{groupgray}  &  & IoU & 0.50 & [0.50, 1.00] & \low & \high \\*
\rowcolor{groupgray}  & \multirow{-5}{*}{BBox} & PG & 0.15 & [0.15, 1.00] & \high & \high \\
\rowcolor{groupgray}  &  & BER & 0.75 & [0.75, 1.00] & \medium & \high \\*
\rowcolor{groupgray}  &  & EPG & 0.75 & [0.75, 1.00] & \medium & \high \\*
\rowcolor{groupgray}  &  & HD & 0.90 & [0.90, 1.00] & \low & \medium \\*
\rowcolor{groupgray}  &  & IoU & 0.75 & [0.75, 1.00] & \medium & \high \\*
\rowcolor{groupgray}  & \multirow{-5}{*}{Region} & PG & 0.01 & [0.01, 1.00] & \high & \high \\
\rowcolor{groupgray}  &  & BER & 0.90 & [0.90, 1.00] & \medium & \medium \\*
\rowcolor{groupgray}  &  & EPG & 0.90 & [0.90, 1.00] & \medium & \medium \\*
\rowcolor{groupgray}  &  & HD & 0.90 & [0.90, 1.00] & \low & \medium \\*
\rowcolor{groupgray}  &  & IoU & 0.75 & [0.75, 1.00] & \medium & \high \\*
\rowcolor{groupgray}  & \multirow{-5}{*}{Boundary} & PG & 0.01 & [0.01, 1.00] & \high & \high \\
\rowcolor{groupgray}  &  & BER & 0.90 & [0.90, 1.00] & \high & \medium \\*
\rowcolor{groupgray}  &  & EPG & 0.90 & [0.90, 1.00] & \high & \medium \\*
\rowcolor{groupgray}  &  & HD & 0.90 & [0.90, 1.00] & \low & \medium \\*
\rowcolor{groupgray}  &  & IoU & 0.90 & [0.90, 1.00] & \high & \medium \\*
\rowcolor{groupgray} \multirow{-20}{*}{GC} & \multirow{-5}{*}{Point} & PG & 0.01 & [0.01, 1.00] & \high & \high \\
\midrule

 &  & BER & 0.75 & [0.75, 1.00] & \low & \high \\*
 &  & EPG & 0.75 & [0.75, 1.00] & \low & \high \\*
 &  & HD & 0.90 & [0.90, 1.00] & \low & \medium \\*
 &  & IoU & 0.75 & [0.75, 1.00] & \medium & \high \\*
 & \multirow{-5}{*}{BBox} & PG & 0.01 & [0.01, 1.00] & \high & \high \\
 &  & BER & 0.90 & [0.90, 1.00] & \medium & \medium \\*
 &  & EPG & 0.90 & [0.90, 1.00] & \medium & \medium \\*
 &  & HD & 0.90 & [0.90, 1.00] & \low & \medium \\*
 &  & IoU & 0.90 & [0.90, 1.00] & \medium & \medium \\*
 & \multirow{-5}{*}{Region} & PG & 0.01 & [0.01, 1.00] & \high & \high \\
 &  & BER & 0.90 & [0.90, 1.00] & \medium & \medium \\*
 &  & EPG & 0.90 & [0.90, 1.00] & \medium & \medium \\*
 &  & HD & 0.90 & [0.90, 1.00] & \low & \medium \\*
 &  & IoU & 0.90 & [0.90, 1.00] & \medium & \medium \\*
 & \multirow{-5}{*}{Boundary} & PG & 0.01 & [0.01, 1.00] & \high & \high \\
 &  & BER & 0.90 & [0.90, 1.00] & \high & \medium \\*
 &  & EPG & 0.90 & [0.90, 1.00] & \high & \medium \\*
 &  & HD & 0.90 & [0.90, 1.00] & \low & \medium \\*
 &  & IoU & 0.90 & [0.90, 1.00] & \high & \medium \\*
\multirow{-20}{*}{GGC} & \multirow{-5}{*}{Point} & PG & 0.01 & [0.01, 1.00] & \high & \high \\
\midrule

\rowcolor{groupgray}  &  & BER & -- & -- & \low & \low \\*
\rowcolor{groupgray}  &  & EPG & -- & -- & \low & \low \\*
\rowcolor{groupgray}  &  & HD & -- & -- & \low & \low \\*
\rowcolor{groupgray}  &  & IoU & -- & -- & \low & \low \\*
\rowcolor{groupgray}  & \multirow{-5}{*}{BBox} & PG & 0.01 & [0.01, 1.00] & \high & \high \\
\rowcolor{groupgray}  &  & BER & -- & -- & \medium & \low \\*
\rowcolor{groupgray}  &  & EPG & -- & -- & \medium & \low \\*
\rowcolor{groupgray}  &  & HD & -- & -- & \low & \low \\*
\rowcolor{groupgray}  &  & IoU & -- & -- & \medium & \low \\*
\rowcolor{groupgray}  & \multirow{-5}{*}{Region} & PG & 0.01 & [0.01, 1.00] & \high & \high \\
\rowcolor{groupgray}  &  & BER & -- & -- & \medium & \low \\*
\rowcolor{groupgray}  &  & EPG & -- & -- & \medium & \low \\*
\rowcolor{groupgray}  &  & HD & -- & -- & \low & \low \\*
\rowcolor{groupgray}  &  & IoU & -- & -- & \medium & \low \\*
\rowcolor{groupgray}  & \multirow{-5}{*}{Boundary} & PG & 0.01 & [0.01, 1.00] & \high & \high \\
\rowcolor{groupgray}  &  & BER & -- & -- & \high & \low \\*
\rowcolor{groupgray}  &  & EPG & -- & -- & \high & \low \\*
\rowcolor{groupgray}  &  & HD & -- & -- & \low & \low \\*
\rowcolor{groupgray}  &  & IoU & -- & -- & \high & \low \\*
\rowcolor{groupgray} \multirow{-20}{*}{IG} & \multirow{-5}{*}{Point} & PG & 0.01 & [0.01, 1.00] & \high & \high \\
\midrule

 &  & BER & 0.90 & [0.90, 1.00] & \low & \medium \\*
 &  & EPG & 0.90 & [0.90, 1.00] & \low & \medium \\*
 &  & HD & -- & -- & \low & \low \\*
 &  & IoU & -- & -- & \low & \low \\*
 & \multirow{-5}{*}{BBox} & PG & 0.01 & [0.01, 1.00] & \high & \high \\
 &  & BER & -- & -- & \medium & \low \\*
 &  & EPG & -- & -- & \medium & \low \\*
 &  & HD & -- & -- & \low & \low \\*
 &  & IoU & -- & -- & \medium & \low \\*
 & \multirow{-5}{*}{Region} & PG & 0.01 & [0.01, 1.00] & \high & \high \\
 &  & BER & -- & -- & \medium & \low \\*
 &  & EPG & -- & -- & \medium & \low \\*
 &  & HD & -- & -- & \low & \low \\*
 &  & IoU & -- & -- & \medium & \low \\*
 & \multirow{-5}{*}{Boundary} & PG & 0.01 & [0.01, 1.00] & \high & \high \\
 &  & BER & 0.90 & [0.90, 1.00] & \high & \medium \\*
 &  & EPG & 0.90 & [0.90, 1.00] & \medium & \medium \\*
 &  & HD & -- & -- & \low & \low \\*
 &  & IoU & 0.90 & [0.90, 1.00] & \high & \medium \\*
\multirow{-20}{*}{SL} & \multirow{-5}{*}{Point} & PG & 0.01 & [0.01, 1.00] & \high & \high \\
\end{longtable}



\end{document}